\documentclass[journal]{IEEEtran}

\pdfoutput=1
\usepackage{graphicx}
\usepackage{subfig}
\usepackage{booktabs,array}
\usepackage{amsmath,amssymb}
\usepackage{mathtools}
\usepackage{siunitx}
\usepackage{bmpsize}
\usepackage{adjustbox}
\usepackage[abs]{overpic}
\usepackage{multirow}
\usepackage{balance}

\usepackage{algorithm}
\usepackage{algpseudocode}
\makeatletter
\def\BState{\State\hskip-\ALG@thistlm}
\makeatother

\algnewcommand{\Initialize}[1]{%
	\Ensure \textbf{Initialize}
}

\usepackage{lmodern}

\renewcommand\thesection{\arabic{section}}
\renewcommand\thesubsection{\thesection.\arabic{subsection}}
\renewcommand\thesubsubsection{\thesubsection.\arabic{subsubsection}}

\renewcommand\thesectiondis{\arabic{section}}
\renewcommand\thesubsectiondis{\thesectiondis.\arabic{subsection}}
\renewcommand\thesubsubsectiondis{\thesubsectiondis.\arabic{subsubsection}}

\usepackage[round,authoryear]{natbib}
\usepackage[usenames,dvipsnames]{color}
\usepackage[colorlinks=true,linkcolor=NavyBlue, citecolor=NavyBlue, urlcolor=NavyBlue]{hyperref}

\ifCLASSINFOpdf
\else
\fi
\begin{document}
%
\title{MedGAN: Medical Image Translation using GANs}
%
%
%

\author{Karim~Armanious\textsuperscript{1,2},
        Chenming~Jiang\textsuperscript{1},
        Marc~Fischer\textsuperscript{1,2},
        Thomas~K\"ustner\textsuperscript{1,2,3},
        Konstantin~Nikolaou\textsuperscript{2},
        Sergios~Gatidis\textsuperscript{2},
        and~Bin~Yang\textsuperscript{1}~\IEEEmembership{\\
        \bigbreak
        \textsuperscript{1}University~of~Stuttgart~,~Institute~of~Signal~Processing~and~System~Theory,~Stuttgart,~Germany\\
        \textsuperscript{2}University~of~T\"ubingen,~Department~of~Radiology,~T\"ubingen,~Germany\\
    	\textsuperscript{3}King's~College~London,~Biomedical~Engineering~Department,~London,~England}
\thanks{\textbf{Corresponding author:} Karim Armanious} \thanks{\textbf{E-mail address:} karim.armanious@iss.uni-stuttgart.de} \thanks{\textbf{Postal address:} Pfaffenwaldring 47 ,70569, Stuttgart, Germany}}
\maketitle

\begin{abstract}

Image-to-image translation is considered a new frontier in the field of medical image analysis, with numerous potential applications. However, a large portion of recent approaches offers individualized solutions based on specialized task-specific architectures or require refinement through non-end-to-end training. In this paper, we propose a new framework, named MedGAN, for medical image-to-image translation which operates on the image level in an end-to-end manner. MedGAN builds upon recent advances in the field of generative adversarial networks (GANs) by merging the adversarial framework with a new combination of non-adversarial losses. We utilize a discriminator network as a trainable feature extractor which penalizes the discrepancy between the translated medical images and the desired modalities. Moreover, style-transfer losses are utilized to match the textures and fine-structures of the desired target images to the translated images. Additionally, we present a new generator architecture, titled CasNet, which enhances the sharpness of the translated medical outputs through progressive refinement via encoder-decoder pairs. Without any application-specific modifications, we apply MedGAN on three different tasks: PET-CT translation, correction of MR motion artefacts and PET image denoising. Perceptual analysis by radiologists and quantitative evaluations illustrate that the MedGAN outperforms other existing translation approaches.

\end{abstract} 

\begin{IEEEkeywords}
Generative adversarial networks, Deep neural networks, Image translation, PET attenuation correction, MR motion correction.
\end{IEEEkeywords}

%
\IEEEpeerreviewmaketitle

\section{Introduction}
%
%
%
%

\IEEEPARstart{I}{n} the field of medical imaging, a wide range of methods is used to obtain spatially resolved information about organs and tissues in-vivo. This includes plain radiography, computed tomography (CT), magnetic resonance imaging (MRI) and positron emission tomography (PET). The underlying physical principles are manifold, producing imaging data of different dimensionality and of varying contrasts. This variety offers various diagnostic options but also poses a challenge when it comes to translation of image information between different modalities or different acquisitions within one modality.

Often, a situation occurs where two imaging modalities or image contrasts provide supplementary information so that two or more acquisitions are necessary for a complete diagnostic procedure. One example is hybrid imaging, e.g. PET/CT where CT is used for the technical purpose of attenuation correction (AC) of PET data \citep{1}. Similarly, CT is used for dosimetry in radiation oncology and has to be acquired in addition to a diagnostic planning MR \citep{2}.

Additionally, optimization of image quality is an important step prior to the extraction of diagnostic information. Especially when using automated image analysis tools, high image quality is required for the accuracy and reliability of the results. In specific situations, the generation of additional image information may be feasible without additional examinations using information from already acquired data. Therefore, a framework which is capable of translating between medical image modalities would shorten the diagnostic procedure by making additional scans unnecessary. This enhanced diagnostic efficiency could prove to be beneficial not only for medical professionals but it also would be more convenient and efficient for patients alike.

Nevertheless, the task of translating from an input image modality to an output modality is challenging due to the possibility of introducing unrealistic information. This would evidently render the synthetic image unreliable for use in diagnostic purposes. However, in specific technical situations, it is not the detailed image content in the synthetic image that is required but rather a global contrast property. In these situations, the translated images are used to enhance the quality of further post-processing tasks rather than diagnosis. An example is PET to CT translation, where the synthetic CT images are not used directly for diagnosis but rather for PET AC.

\subsection{Related work}
\bigbreak
In the last decade, several computational methods have been introduced for the translation of medical images using machine learning approaches. For example, structured random forest was used in conjunction with an auto-context model to iteratively translate MR patches into corresponding CT for the purpose of PET AC \citep{Huynh2016EstimatingCI}. For a given MR image, the synthetic CT patches are combined to give the final AC prediction. Going in a similar direction, pseudo-CT images were predicted from input MR patches using a k-nearest neighbour (KNN) regression algorithm. The efficiency of the prediction was first improved by local descriptors learned through a supervised descriptor learning (SDL) algorithm \citep{7493373} and more recently through the combination of feature matching with learned non-linear local descriptors \citep{8249878}. In another application domain, the correction of rigid and non-rigid motion artefacts in medical images could be viewed as a domain translation problem from motion-corrupted images into motion-free images. K\"ustner et al \citep{Kustner2017MRbasedRA} presented a method for cardiac and respiratory motion correction for PET images via simultaneously acquired MR motion model and a corresponding compressed sensing reconstruction scheme. A further interesting application is resolution enhancement of MR images. A super-resolution method, which translates a low-resolution MR image into a higher-resolution version, was  developed based on a sparse representation framework which incorporates multi-scale edge analysis and a dimensionality reduction scheme for more efficient reconstruction \citep{article}. 


Recently, the computer vision community has gained momentum in the area of medical image analysis \citep{Litjens2017ASO}. This is due to recent advances in a range of applications such as lesion detection and classification \citep{7404017,7576695}, semantic segmentation \citep{5435435,4324242}, registration \citep{7393571} and image enhancement \citep{7547547547,Bahrami2016ConvolutionalNN,Oktay2016MultiinputCI} with the development of deep learning algorithms, especially the convolutional neural network (CNN) \citep{543254345}. This has led to the development of several approaches for the generation and translation of image data. The most prominent of those are GANs.

In 2014, I. Goodfellow \citep{Goodfellow2014GenerativeAN} introduced Generative Adversarial Networks (GANs). They are generative models with the objective of learning the underlying distribution of training data in order to generate new realistic data samples which are indistinguishable from the input dataset. Prior to the introduction of GANs, state-of-the-art generation models, such as Variational Autoencoders (VAE) \citep{523452345,pmlr-v32-rezende14}, tackled this task by performing explicit density estimation. GANs constitute an alternative to this by defining a high-level goal such as "generate output data samples which are indistinguishable from input data" and minimizing the loss function through a second adversarial network instead of explicitly defining it.

The main underlying principle of GANs is that of rivalry and competition between two co-existing networks. The first network, the generator, takes random noise as input and outputs synthetic data samples. The second network, the discriminator, acts as a binary classifier which attempts to distinguish between real training data samples and fake synthetic samples from the generator. In the training procedure, the two networks are trained simultaneously with opposing goals. The generator is instructed to maximize the probability of fooling the discriminator into thinking the synthetic data samples are realistic. On the other hand, the discriminator is trained to minimize the cross entropy loss between real and generated samples, thus maximize the probability of correctly classifying real and synthetic images. Convergence is achieved by GANs from a game theory point of view by reaching Nash equilibrium \citep{DBLP:journals/corr/ZhaoML16}. Thus, the distribution of the generator network will converge to that of the training data and the discriminator will be maximally confused in distinguishing between real and fake data samples.

Researchers have adapted adversarial networks for different tasks. Intuitively, GANs are the state-of-the-art model for image synthesis with recent models achieving unprecedented levels of image realism \citep{DBLP:journals/corr/abs-1710-10196}. C. Ledig \citep{DBLP:journals/corr/LedigTHCATTWS16} achieved state-of-the-art results in the field of image super-resolution via the combination of the adversarial loss together with content loss in the Super-Resolution GAN (SR-GAN) framework. Other applications utilizing the GANs includes classification \citep{Salimans2016ImprovedTF}, image denoising \citep{DBLP:journals/corr/ZhangSP17} and text to image synthesis \citep{DBLP:journals/corr/ZhangXLZHWM16} among many others. The most relevant utilization of GAN in the field of medical image analysis is image-to-image translation.

In 2016, P. Isola \citep{DBLP:journals/corr/IsolaZZE16} introduced the pix2pix GAN framework as general solution to supervised image-to-image translation problems. In this case, the generator receives as input an image from the input domain (e.g a grayscale photo) and is tasked to translate it to the target domain (e.g a coloured photo) by minimizing a pixel-reconstruction error (L1 loss) as well as the adversarial loss. On the other hand, the discriminator is tasked to differentiate between the fake output of the generator and the desired ground truth output image. Several modifications of this framework have been developed to enhance the quality of the output images. For example, PAN \citep{8358814} replaced the pixel loss with a feature matching loss from the discriminator to reduce the blurriness of the output images \citep{final} . For the purpose of one-to-many translation, Fila-sGAN \citep{Zhao2017SynthesizingFS} utilized a pre-trained network for the calculation of style losses \citep{DBLP:journals/corr/JohnsonAL16} to transfer the texture of input style images onto the translated image. Moreover, several unsupervised variants were introduced that do not require a dataset of paired input/target images for training, such as Cycle-GANs \citep{DBLP:journals/corr/ZhuPIE17} and Disco-GANs \citep{DBLP:journals/corr/KimCKLK17}. 

Recently, GANs have been gaining more attention in the medical field especially for image-to-image translation tasks. For instance, a pix2pix architecture with an added gradient-based loss function was utilized for the translation from MR to CT images \citep{8310638}. This architecture suffered from limited modelling capacity due to patch-wise training. This rendered end-to-end training infeasible. Instead, it is necessary to train several GAN frameworks one after another via an auto-context model to refine the results. A similar but unsupervised approach was proposed in \citep{DBLP:journals/corr/abs-1708-01155} via Cycle-GANs. Pix2pix GANs were also utilized for the task of denoising low dose CT images by translating it into a high dose counterpart \citep{45324}. Also for the task of CT denoising, \citep{8340157} utilized a pre-trained network for the calculation of feature matching losses together with the Wasserstein distance loss. Synonymous with the above mentioned work, \citep{8233175} utilized a largely similar architecture for the task of compressed sensing (CS) MRI reconstruction. 

Most relevant to our work, \citep{8327637} presented a generator architecture specifically tailored for the task of CS MRI reconstruction. The architecture consists of two residual networks concatenated in an end-to-end manner. Although the results of such an architecture surpassed that of conventional pix2pix, it suffers from the limitation of being specific to CS MRI reconstruction and not extendable to other translation tasks in which the target domain differs significantly from the input domain (e.g. MR to CT translation).  Other medical translation tasks have also been recently explored such as CT to PET \citep{DBLP:journals/corr/abs-1802-07846} and 2T MR to 1T MR translation \citep{Yang2018MRIIT, DBLP:journals/corr/abs-1802-01221}. The above-presented approaches is an overview of the utilization of  GANs for medical image-to-image translation tasks. However, new applications domains are continually being explored by radiologists and engineers alike. A more in-depth survey of utilization of adversarial networks for medical imaging can be found in \citep{reviewpaper}.

\subsection{Contributions}


An analysis of the above mentioned medical adversarial frameworks identifies a common phenomenon. A large portion of the existing approaches are application-specific or suffer from a limited modelling capacity. Thus, these models can not be easily re-applied to other medical imaging tasks.

In this work, we propose a new GAN framework for medical image-to-image translation, titled MedGAN. Inspired by previous works such as ResNets \citep{7780459}, pix2pix, PAN and Fila-sGAN, our work combines the fragmented benefits of previous translation approaches with a new high-capacity generator architecture. The resultant framework is applicable to different medical tasks without application-specific modifications. Rather than diagnosis, the main purpose of MedGAN is to enhance further technical post-processing tasks that require globally consistent image properties. The proposed MedGAN framework outperforms other existing translation approaches in qualitative and quantitative comparisons.\\

\begin{figure*}[!ht]
	\centering
	
	\includegraphics[width=1\textwidth]{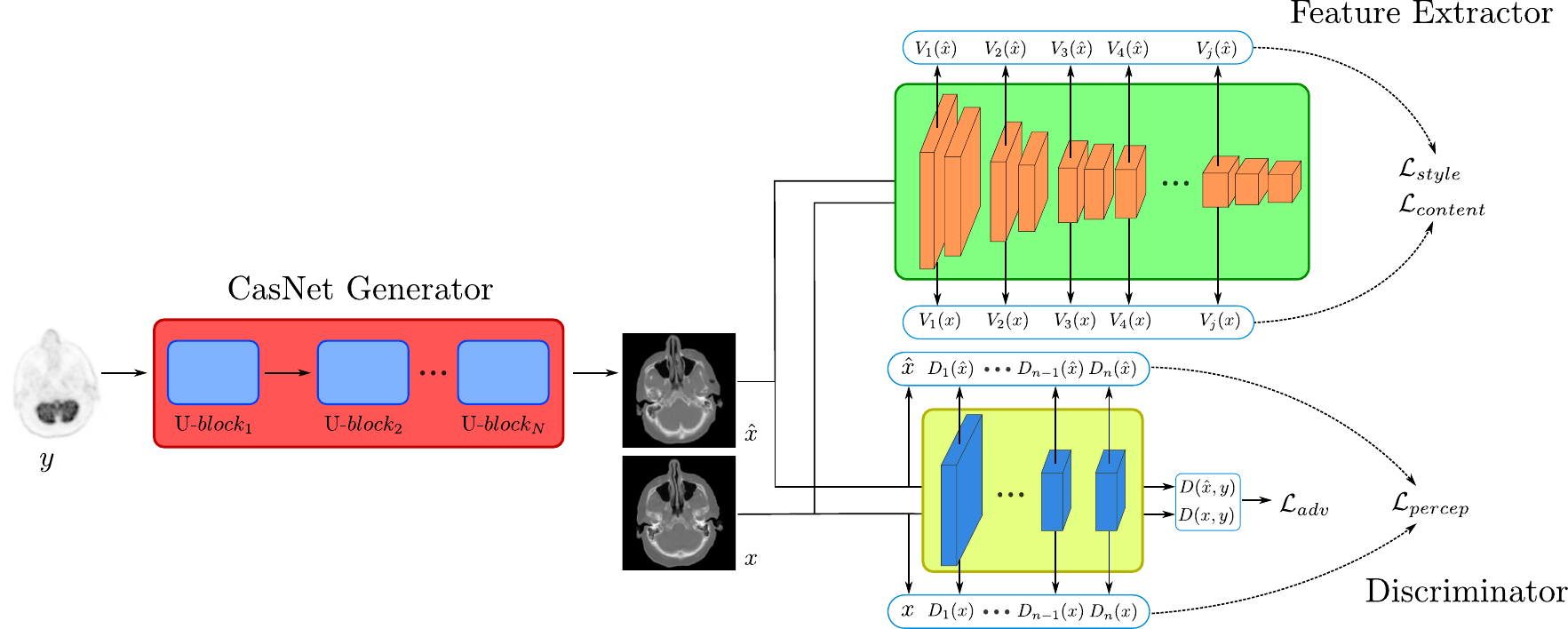}
	
	\caption{Overview of the MedGAN framework comprising of a novel CasNet generator architecture, a discriminator and a pre-trained feature extractor. The generator $G$ is tasked with translating input images from the source domain $y$ (e.g PET) to the target domain $\hat{x}$ (e.g. CT) through progressive refinement via encoder-decoder blocks. The adversarial discriminator $D$ is trained to distinguish between real and transformed images and co-serve as a trainable feature extractor to calculate the modified perceptual loss. The pre-trained feature extractor is used to extract deep rich features $V_i(\hat{x})$ to calculate style transfer losses in order for the output to match the target's style, textures and content.}
	
	\label{1}
\end{figure*}

Our contributions are summarized as follows:

\begin{itemize}
	\item MedGAN as a new framework for medical translation tasks. MedGAN captures the high and low frequency components of the desired target modality by combining the adversarial framework with a new combination of non-adversarial losses. Specifically, a modified perceptual loss is utilized together with style-transfer losses.
	
	\item A new generator architecture, named CasNet. Inspired by ResNets, this architecture chains together several fully convolutional encoder-decoder networks with skip connections into a single generator network. As the input medical image propagates through the encoder-decoder pairs, the translated images will progressively be refined to ensure a high resolution and crisp output. CasNet is an end-to-end architecture not specific to any particular application. Concurrent to this work, a similar architecture, refereed to as stacked U-Nets, was developed for natural image segmentation \citep{19} .
	
	\item Application of MedGAN on three challenging tasks in medical imaging with no application-specific modifications to the hyperparameters. These are translation from PET images into synthetic CT images, PET image denoising and finally the retrospective correction of rigid MR motion artefacts.
	
	\item Quantitative and qualitative comparison of MedGAN with other adversarial translation frameworks. A further analysis of individual loss functions was performed to illustrate that MedGAN is more than the sum of its components.
	
	\item The subjective performance and fidelity of the translated medical images was investigated from a medical perspective. This was done by conducting a perceptual study in which 5 experienced radiologists where tasked to rate the results.
\end{itemize}

\section{Materials and methods}

An overview of the proposed MedGAN framework for medical image-to-image translation tasks is presented in \hyperref[1]{Fig.~\ref*{1}}. In this section, the different loss components and network architecture of MedGAN will be presented starting first with some preliminary information.

\subsection{Preliminaries}

\subsubsection{Generative adversarial networks}
GANs consist of two main components, a generator and a discriminator. The generator $G$ receives as input samples $z$ from a prior noise distribution $p_{\small\textrm{noise}}$ (e.g. a normal distribution) and is tasked to map it to the data space $\hat{x} = G(z)$ inducing a model distribution $p_{\small\textrm{model}}$. On the other hand, the discriminator is a binary classifier whose objective is to classify data samples $x \sim p_{\small\textrm{data}}$ as real, $D(x) = 1$, and generated samples $\hat{x} \sim p_{\small\textrm{model}}$ as fake, $D(\hat{x}) = 0$.

Both networks are pitted in a competition against each other. The generator attempts to produce samples which are indistinguishable from the real samples, $p_{\small\textrm{model}} \approx p_{\small\textrm{data}}$, thus fooling the discriminator. In the meantime, the discriminator's objective is to avoid being fooled through learning meaningful features which better distinguish between real and generated samples. This concept of adversary between opposing networks is well represented by the principles of game theory via the following min-max optimization task:
\begin{equation}
\min_{G} \max_{D}  \mathcal{L}_{\small\textrm{GAN}}
\end{equation}
where $\mathcal{L}_{\small\textrm{GAN}}$ is the adversarial loss given by: 
\begin{equation}
	\mathcal{L}_{\small\textrm{GAN}}= \mathbb{E}_{x \sim p_{\small\textrm{data}}} \left[\textrm{log} D(x) \right] + \mathbb{E}_{z \sim p_{\small\textrm{noise}}} \left[\textrm{log} \left( 1 - D\left(G(z)\right) \right) \right]
\end{equation}

The cost function of each network is dependent on the opposing network parameters, therefore convergence is achieved by reaching Nash equilibrium (i.e. saddle point) rather than a local minimum. The theoretically motivated approach of training the discriminator to optimality for a fixed generator network typically results in a vanishing gradient problem. Alternatively, it was found that alternating between updating the opposing networks one at a time while fixing the other helps to avoid this problem \citep{Goodfellow2014GenerativeAN}.\\

\subsubsection{Image-to-image translation}
The underlying principle of adapting adversarial networks from image generation to translational tasks is replacing the generator network by its conditional variant (cGAN) \citep{DBLP:journals/corr/IsolaZZE16}. In this case, the generator aims to map a source domain image $y \sim p_{\small\textrm{source}}$ into its corresponding ground truth target image $x \sim p_{\small\textrm{target}}$ via the mapping function $G(y,z) = \hat{x} \sim p_{\small\textrm{model}}$. This can generally be viewed as a regression task between two domains that share the same underlying structures but differ in surface appearance. An example would be the translation of grayscale imagery to corresponding colour imagery. However, instead of using manually constructed loss functions to measure the similarity between the translated and target images, cGAN utilizes a binary classifier, the discriminator, as an alternative. 

In this case, the adversarial loss is rewritten as:
\begin{equation}
	\mathcal{L}_{\small\textrm{cGAN}} = \mathbb{E}_{x,y} \left[\textrm{log} D(x,y) \right] + \mathbb{E}_{z,y} \left[\textrm{log} \left( 1 - D\left(G(y,z),y\right) \right) \right]
	\label{e3}
\end{equation}
such that the discriminator aims to classify the concatenation of the source image $y$ and its corresponding ground truth image $x$ as real, $D(x,y) = 1$ , while classifying $y$ and the transformed image $\hat{x}$ as fake, $D(\hat{x},y) = 0$.

Nevertheless, image-to-image translation frameworks that rely solely on the adversarial loss function do not produce consistent results. More specifically, the output images may not share a similar global structure as the desired ground truth image. To counteract this issue, a pixel reconstruction loss, such as the L1 loss, is usually incorporated \citep{DBLP:journals/corr/IsolaZZE16,Zhao2017SynthesizingFS}. This is achieved by calculating the mean absolute error (MAE) between the target and synthetic images:

\begin{equation}
	\mathcal{L}_{\small\textrm{L1}} = \mathbb{E}_{x,y,z} \left[\lVert{x - G(y,z)}\rVert_1\right]
\end{equation}
such that the final training objective is given by:
\begin{equation}
\min_{G} \max_{D}  \mathcal{L}_{\small\textrm{cGAN}} + \lambda \mathcal{L}_{\small\textrm{L1}}
\end{equation}
with $\lambda > 0$ as a weighting hyperparameter. 

\subsection{Perceptual loss}

Despite the advantages of pixel-reconstruction losses, they also commonly lead to blurry results \citep{DBLP:journals/corr/PathakKDDE16,12312312}. As a result, the translation frameworks which utilize such loss functions often result in outputs with well maintained global structures at the cost of distortions and loss of details.  Such pixel losses fail to capture the perceptual quality of human judgement. This is easily examined when inspecting two identical images shifted by a few pixels from each other. Unlike the human brain which will immediately capture the similarities between the images, a pixel-wise comparison will judge the images as vastly different \citep{DBLP:journals/corr/JohnsonAL16}.
This phenomenon is critical in the domain of medical images where small structures could significantly alter the diagnostic information of an image. 


Therefore, to capture the discrepancy between the high frequency components within an image a perceptual loss is additionally utilized. This loss is based on using the discriminator network a trainable feature extractor to extract intermediate feature representations. The MAE between the feature maps of the target images $x$ and the translated images $\hat{x}$ is then calculates as:

%

\begin{equation}
	P_i \left(G(y,z),x\right) = \frac{1}{h_i w_i d_i}  \lVert{D_i\left(G(y,z),y\right) - D_i\left(x,y\right)}\rVert_1
\end{equation}
where $D_i$ denotes the feature representations extracted from the $i^{\textrm{th}}$ hidden layer of the discriminator network, and $h_i$, $w_i$ and $d_i$ represents the height, width and depth of the feature space, respectively.

The perceptual loss is then be formulated as:

\begin{equation}
	\mathcal{L}_{\small\textrm{perceptual}} = \sum_{i = 0}^{L} \lambda_{pi} P_i \left(G(y,z),x\right)
	\label{7}
\end{equation}
with $L$ the number of hidden layers of the discriminator and $\lambda_{pi} > 0$ is a tuning hyperparameter which represents the influence of the $i^{\textrm{th}}$ layer.  Analogous to $\lambda$ for the L1 loss, $\lambda_{pi}$ is optimized prior to the training of the network for each layer $i$ via a hyperparameter optimization process. This will be further discussed in the end of this section.
%

It is important to note that unlike other GAN frameworks which utilize feature matching losses (e.g. PAN \citep{8358814}), the proposed perceptual loss does not eliminate the pixel reconstruction component. This is due to the observation that penalizing the discrepancy in the pixel-space has a positive impact on the quality of the results and should not be ignored for the sake of strengthening the output details.

\begin{figure*}[!t]
	\centering
	
	\includegraphics[width=1\textwidth]{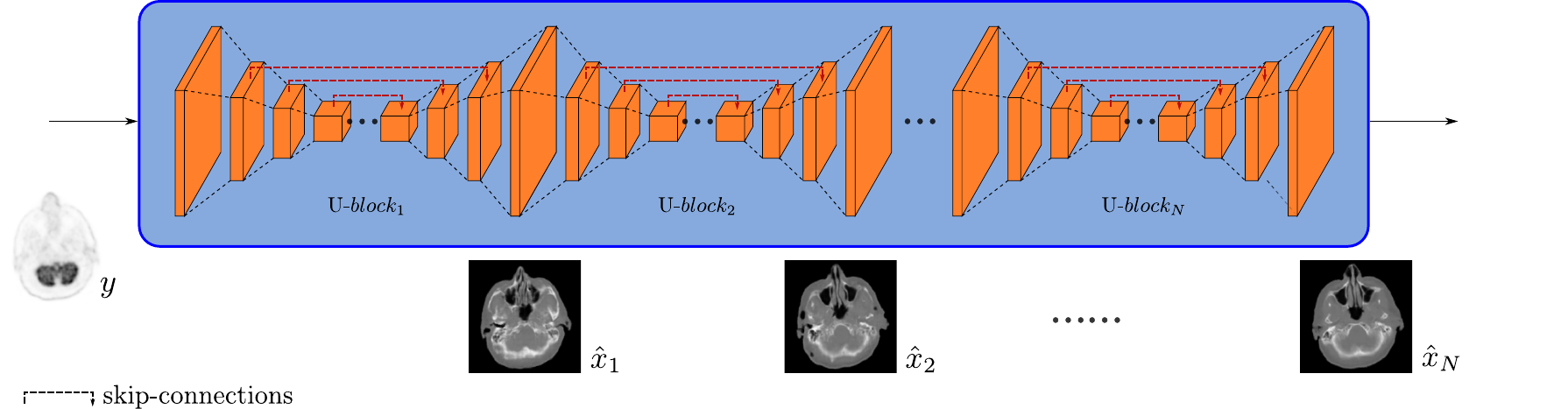}
	
	\caption{The proposed CasNet generator architecture. CasNet concatenates several encoder-decoder pairs (U-blocks) to progressively refine the desired output image.}
	\label{2}
\end{figure*}

Additionally, in order to extract more meaningful features for the calculation of the perceptual loss, it is necessary to stabilize the adversarial training of the discriminator. For this purpose, spectral normalization regularization was utilized \citep{DBLP:journals/corr/abs-1802-05957}. This is achieved by normalizing the weight matrix $\theta_{D,i}$ of each layer $i$ in the discriminator:

\begin{equation}
	\theta_{D,i} = \theta_{D,i} / \delta(\theta_{D,i})
\end{equation}
where $\delta(\theta_{D,i})$ represents the spectral norm of the matrix $\theta_{D,i}$. As a result, the Lipschitz constant of the discriminator function $D\left(x,y\right)$ will be constrained to $1$. Practically, instead of applying singular value decomposition for the calculation of the spectral norm, an approximation via the power iteration method $\hat{\delta}(W_i)$ was used instead in order to reduce the required computation complexity \citep{DBLP:journals/corr/abs-1802-05957}.
\subsection{Style transfer losses}
Image translation of medical images is a challenging task since both global fidelity and high frequency sharpness, and thus clarity of details, are required for further medical post-processing tasks. For example, in PET to CT translation, the synthesized CT image must exhibit detailed bone structure for accurate PET attenuation correction. Furthermore, in the correction of MR motion artefacts, the resulting image must contain accurate soft-tissue structures as this will affect the results of subsequent post-processing tasks such as segmentation and organ volume calculation.

To achieve the required level of details, MedGAN incorporates non-adversarial losses from recent image style transfer techniques \citep{7780634,DBLP:journals/corr/JohnsonAL16}. These losses transfer the style of an input image onto the output image, matching their textures and details in the process. Similar to the perceptual loss, features from the hidden layers of a deep CNN are used for loss calculations. However, instead of utilizing the discriminator, a feature extractor, pre-trained for an image classification task, is used. Compared to the discriminator, the pre-trained network has the advantage of being a deeper neural network consisting of multiple convolutional blocks. This allows the extraction of rich features from a larger receptive field to also enhance the global structures of the translated images in addition to the fine details. Style transfer losses can be divided into two main components: style loss and content loss. 

\subsubsection{Style loss}

The style loss is used to penalize the discrepancy in the style representations between the translated images and their corresponding target images. The style distribution can be captured by calculating the correlations between feature representations in the spatial extent. $V_{j,i}(x)$ denote the feature maps extracted from the $j^{\textrm{th}}$ convolutional block and $i^{\textrm{th}}$ layer of the  feature extractor network for input image $x$. The feature maps have then the size $h_j \times w_j \times d_j$ with $h_j$, $w_j$, $d_j$ being the height, width and spatial depth, respectively. In this work, only the first layer of each convolutional block is used, thus the sub-index $i$ is assumed to be 1 and will be omitted in the following notations. The feature correlations are represented by the Gram matrix $Gr_j(x)$ of each convolutional block. This matrix is of the shape $d_j \times d_j$ and its elements are calculated by the inner product between feature maps over the height and width dimensions:
\begin{equation}
	Gr_j(x)_{m,n} = \frac{1}{h_j w_j d_j} \sum_{h = 1}^{h_j} \sum_{w = 1}^{w_j} V_{j}(x)_{h,w,m} V_{j}(x)_{h,w,n}
\end{equation}

The style loss is then calculated as the Frobenius squared norm of the differences between the feature correlations of the translated outputs $\hat{x}$ and the ground truth inputs $x$:
 
\begin{equation}
\mathcal{L}_{\small\textrm{style}} = \sum_{j = 1}^{B} \lambda_{sj} \frac{1}{4 d_j^2} \lVert{Gr_j\left(G(y,z)\right) - Gr_j\left(x\right)}\rVert_F^{2}
\end{equation}
 where $\lambda_{sj} > 0$ is also a tuning hyperparameters representing the weight of the contribution of the $j^{\textrm{th}}$ convolutional block and $B$ is the total number of convolutional blocks.\\
 
\subsubsection{Content loss}
The content loss directly penalizes the differences between feature representations extracted from the feature extractor network. Contrary to the style loss, the content loss does not capture discrepancies in style or texture. However, it serves an auxiliary purpose analogous to that of the pixel-reconstruction loss by enhancing low frequency components and ensuring global consistency of the transformed images. The content loss is given by:

\begin{equation}
\mathcal{L}_{\small\textrm{content}} = \sum_{j = 1}^{B} \lambda_{cj} \frac{1}{h_j w_j d_j} \lVert{V_j\left(G(y,z)\right) - V_j\left(x\right)}\rVert_F^{2}
\end{equation}\\
where $\lambda_{cj} > 0$ is a hyperparameter representing the influence of the first layer of the $j^{\textrm{th}}$ convolutional block.
\subsection{MedGAN architecture}
\subsubsection{U-blocks}

The task of image-to-image translation can be described as mapping a high dimensional input tensor into an output tensor with different surface appearance but of the same underlying structure. From another aspect, the main architectural consideration of the MedGAN framework is robustness to different input modalities with no application-specific modifications. Therefore, the fundamental building block of MedGAN was chosen to be an encoder-decoder architecture, which we refer to as a U-block.

A U-block is a fully convolutional encoder-decoder network following the architecture introduced in \citep{DBLP:journals/corr/IsolaZZE16}. It is inspired by U-nets \citep{10.1007/978-3-319-24574-4_28} which have been adapted according to the architectural guidelines in \citep{34532452} to stabilize the adversarial training process. The encoding path maps the image from the input domain, in $256 \times 256$ resolution, into a high level representation using a stack of 8 convolutional layers each followed by batch normalization and Leaky-ReLU activation functions. The number of convolutional filters is 64, 128, 256, 512, 512, 512, 512 and 512 respectively with kernel size $4 \times 4$ and stride 2. For the desired purpose of medical image translation, stochasticity is not desired and the encoding path only receives the source domain image $y$ as input. The subscript $z$, in \hyperref[e3]{Eq.~(\ref*{e3})}, denoting input noise samples will hence be omitted from future notations. The decoding path mirrors the encoding architecture albeit utilizing fractionally strided deconvolutions. This enlarges the resolution by a factor of two after each layer, which inverts the downsampling by the encoding path and maps from the high level representation into the output image domain. The upsampling path consists of 512, 1024, 1024, 1024, 1024, 512, 256 and 128 filters, respectively, in each of the layers which utilize ReLU activation functions except for the last deconvolutional layer which uses a Tanh activation instead.

Additionally, a U-block contains skip-connections which concatenate spatial channels between mirrored layers in the encoder and decoder paths, e.g. between the $2^{\textrm{nd}}$ encoding layer and the $7^{\textrm{th}}$ decoding layer. These connections are fundamental for image transformation tasks since they pass critical low level information between the input and output images. This information will otherwise be lost through the bottleneck layer leading to severe degradation in output quality.

\subsubsection{CasNet}

Translation of medical images poses a challenge compared to regular image transformation tasks. This is due to the amount of relevant medical information contained in detailed structures in the images which can be lost or distorted during the translation process. In order to circumvent this issue, current approaches utilize either specialized architectures for a given medical transformation task \citep{8327637} or require the training of several frameworks one after the other \citep{8310638}. In order to construct a non-application-specific solution the CasNet architecture is proposed, illustrated in \hyperref[2]{Fig.~\ref*{2}}.

Inspired by ResNets \citep{7780459}, which cascades the so-called residual blocks, CasNets increases the generative capabilities of MedGAN by concatenating several U-blocks in an end-to-end manner. This is done such that the output of the first U-block is passed as the input of the second block till the $N^{\textrm{th}}$ block. As a result, the translation task is carried out using the collective capacity of the U-blocks in an end-to-end manner. Thus, the translated outputs are progressively refined as they pass through the encoder-decoder pairs. Backpropagation of the loss gradients through such network depth may result in a vanishing gradient problem. However, due to the utilization of skip connections within individual U-blocks this problem is mitigated.

Although CasNets and ResNets share the same basic principle of concatenating a more basic building block, fundamental differences exist between the two networks. The first is concerning the depth. Residual blocks consist of only 2-4 convolutional layers whereas U-blocks have a deeper architecture of 16 convolutional layers, which increases the generative capacity of CasNets. Moreover, CasNets utilize intermediate skip connections to pass low level information and prevent vanishing gradients instead of using identity mappings to connect the input image to the output of the residual block.\\

\subsubsection{Discriminator architecture}

For the discriminator, a modified PatchGAN architecture is utilized, proposed in \citep{DBLP:journals/corr/IsolaZZE16}. Instead of classifying the target and output images as being real or not, PatchGAN is designed to have a reduced receptive field such that it divides the input images convolutionally into smaller image patches before classifying them and averaging out the result. Consequently, the discriminator's attention is restricted to small image patches which encourage high frequency correctness and enables detailed outputs by the generator. Generally, $70 \times 70$ patches is the conventional patch size to utilize in order to avoid the typical tiling artefacts with smaller patch sizes. However, we empirically found that the utilization of smaller patches in combination with the previously introduced non-adversarial losses, e.g. perceptual and style transfer losses, promotes sharper results while eliminating conventional tiling artefacts. As a result, a $16 \times 16$ patch size is utilized by incorporating two convolutional layers with 64 and 128 spatial filters followed by batch normalization and Leaky-ReLU activation functions. Lastly, to output the required confidence probability map, a convolution layer of output dimension 1 and a sigmoid activation function is used.
\begin{figure*}[!t]
	\begin{minipage}[t]{0.32\linewidth}
		\centering
		\begin{overpic}[width=0.430\textwidth]%
			{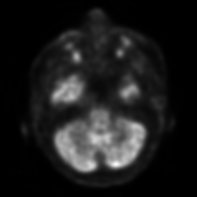}
			\centering
			\put(25,78){Input}
		\end{overpic}
		\begin{overpic}[width=0.430\textwidth]%
			{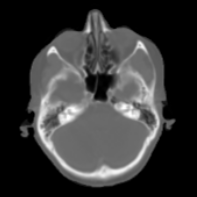}
			\centering
			\put(22,78){Target}
		\end{overpic}		
		
		(a) PET-CT translation
	\end{minipage}%
	\vrule
	\begin{minipage}[t]{0.32\linewidth}
		\centering
		\begin{overpic}[width=0.430\textwidth]%
			{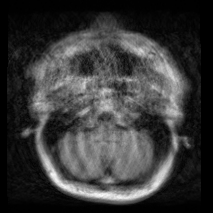}
			\centering
			\put(25,78){Input}
		\end{overpic}
		\begin{overpic}[width=0.430\textwidth]%
			{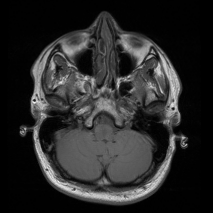}
			\centering
			\put(22,78){Target}
		\end{overpic}
		
		(b) MR motion correction
	\end{minipage}
	\vrule
	\begin{minipage}[t]{0.32\linewidth}
		\centering
		\begin{overpic}[width=0.430\textwidth]%
			{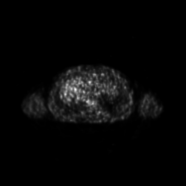}
			\centering
			\put(25,78){Input}
		\end{overpic}
		\begin{overpic}[width=0.430\textwidth]%
			{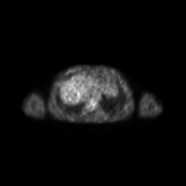}
			\centering
			\put(18,78){Target}
		\end{overpic}
		
		(c) PET denoising
	\end{minipage}
	\caption{An example of the three datasets used for the qualitative and quantitative evaluation of the MedGAN framework.}
	\label{3}
\end{figure*}
\begin{algorithm} [!t]
	\caption{Training pipeline for MedGAN}\label{euclid}
	\label{a1}
	\begin{algorithmic}[1]
		\Require Paired training dataset $\{\left(x_l,y_l\right)\}_{l=1}^T$
		\Require Number of training epochs $N_{\textrm{epoch}} = 200$, number of training iterations for generator $N_\textrm{G} = 3$, $\lambda_1 = 20$ and $\lambda_2 = \lambda_3 = 0.0001$
		\Require Load pretrained VGG-19 network parameters
		\Ensure Weight parameters of generator and discriminator $\theta_G$, $\theta_D$
		\For{$n = 1,...,N_{\textrm{epoch}}$}
		\For{$l = 1,...,T$}
		\For{$t = 1, ..., N_G$}
		\State $\mathcal{L}_{\small\textrm{cGAN}} \gets -\textrm{log} \left(D\left(G(y_l),y_l\right) \right)$
		\State $\mathcal{L}_{\small\textrm{perceptual}} \gets \sum_{i} \lambda_{pi} P_i \left(G(y_l),x_l\right)$
		\State $\mathcal{L}_{\small\textrm{style}} \gets \sum_{j}  \frac{\lambda_{sj}}{4 d_j^2} \lVert{Gr_j\left(G(y_l)\right) - Gr_j\left(x_l\right)}\rVert_F^{2}$
		\State $\mathcal{L}_{\small\textrm{content}} \gets \sum_{j}  \frac{\lambda_{cj}}{h_j w_j d_j} \lVert{V_j\left(G(y_l)\right) - V_j\left(x_l\right)}\rVert_F^{2}$
		\State $\theta_G \overset{+}{\gets} - \nabla_{\theta_G} [\mathcal{L}_{\small\textrm{cGAN}} + \lambda_1 \mathcal{L}_{\small\textrm{perceptual}} + \lambda_2 \mathcal{L}_{\small\textrm{style}}$\newline 
		$\hspace*{30.5mm} + \lambda_3 \mathcal{L}_{\small\textrm{content}}]$ 
		\EndFor
		\State \small $\mathcal{L}_{\small\textrm{cGAN}} \gets \textrm{log} \left(D\left(x_l,y_l\right) \right) + \textrm{log} \left(1-D\left(G(y_l),y_l\right) \right)$
		\State $\theta_D \overset{+}{\gets} \nabla_{\theta_D} \left[\mathcal{L}_{\small\textrm{cGAN}}\right]$
		\State Spectral normalization: $\theta_{D,i} = \theta_{D,i} / \delta(\theta_{D,i})$
		\EndFor
		\EndFor
	\end{algorithmic}
\end{algorithm}
\subsection{MedGAN framework and training}
In summary, the MedGAN framework consists of a CasNet generator network penalized from the perceptual and pixel perspectives via an adversarial discriminator network. Additionally, MedGAN utilizes style transfer losses to ensure that translated output matches the desired target image in style, texture and content. The framework is trained via a min-max optimization task using the following cumulative loss function:
\begin{equation}
	\mathcal{L}_{\small\textrm{MedGAN}} = \mathcal{L}_{\small\textrm{cGAN}} + \lambda_1 \mathcal{L}_{\small\textrm{perceptual}}  + \lambda_2 \mathcal{L}_{\small\textrm{style}} + \lambda_3 \mathcal{L}_{\small\textrm{content}}
\end{equation}
where $\lambda_1$, $\lambda_2$ and $\lambda_3$ are hyperparameters that balance out the contribution of the different loss components. As a result of extensive hyperparameter optimization, $\lambda_1 = 20$ and $\lambda_2 = \lambda_3 = 0.0001$ was utilized. Additionally, $\lambda_{pi}$ was chosen to allow both layers of the discriminator to have equal influence on the loss. Similarly, $\lambda_{cj}$ was set to allow all but the deepest convolutional blocks to influence the content loss. However, the style loss $\lambda_{sj}$ was chosen to include only the influence of the first and last convolutional blocks of the pre-trained VGG-19 network. Regarding the feature extractor, a deep VGG-19 network pre-trained on ImageNet classification task \citep{DBLP:journals/corr/SimonyanZ14a} was used. It consists of 5 convolutional blocks, each of 2-4 layers, and three fully connected layers. Although it is pre-trained on non-medical images, the features extracted by the VGG-19 network was found to be beneficial in representing the texture and style information as will be shown in the following results section. For training, we make use of the ADAM optimizer \citep{DBLP:journals/corr/KingmaB14} with momentum value of $0.5$ and a learning rate of $0.0002$. Instance normalization was applied with a batch size of 1, which was shown to be beneficial for image translation tasks \citep{DBLP:journals/corr/UlyanovVL16}. For the optimization of MedGAN, the patch discriminator was trained once for every three iterations of training the CasNet generator. This leads to a more stable training and produces higher quality results. The entire training process is illustrated in Algorithm 1. 

The MedGAN framework was trained on a single Nvidia Titan-X Gpu with a CasNet generator architecture consisting of $N = 6$ U-blocks. The training time is largely dependent on the size of the dataset used but was found to be an average of 36 hours. The inference time, however, was found to be $115$ milliseconds for each test image. The implementation of the MedGAN framework will be made publicly available upon the publishing of this work\footnote{https://github.com/KarimArmanious}.

\section{Experimental evaluations}

\subsection{Datasets}

 To showcase MedGAN as a non-application-specific framework, MedGAN was directly applied on three challenging tasks in medical imagery. No task-specific modifications to the hyperparameters or architectures was applied. The utilized datasets are illustrated in \hyperref[3]{Fig.~\ref*{3}}.

For the first application, PET images are translated into corresponding synthetic CT images. This is a non-trivial task since the target modality contains more detailed information, e.g. bone structures and soft tissues, compared to the input source modality. For that purpose, an anonymized dataset of 46 patients of the brain region acquired on a joint PET/CT scanner (SOMATOM mCT, Siemens Healthineers, Germany) was used. The CT data has an original resolution of $0.85 \times 0.85 \times 5 \ \textrm{mm}^{3}$ and a matrix size of $512 \times 512$, while PET data have a voxel size of $2 \times 2 \times 3 \ \textrm{mm}^{3}$ and a matrix size of $400 \times 400$. The resolution of both modalities was resampled to a voxel size of $1 \times 1 \times 1 \ \textrm{mm}^{3}$, aligned using the header information and then centre cropped to extract the relevant head region. Due to hardware limitations, only 2-dimensional axial slices of resolution $256 \times 256$ pixels were used during the training process, with a dataset of 1935 paired training images from 38 patients, and 414 images from 8 separate patients for validation.

The second application is concerned with the retrospective correction of motion artefacts in MR images. Motion-corrupted MR images was translated into corresponding motion-free images. This is a challenging task, not only because of the severity of rigid motion artefacts in the acquired datasets but also because of the difficulty achieving pixel-wise alignment between motion free and motion corrupted MR scans taken sequentially in time. This highlights the robustness of MedGAN against alignment errors in the required training datasets. An anonymized dataset of 11 volunteers from the brain region was acquired using a clinical MR scanner (Biograph mMR 3 Tesla, Siemens Healthineers, Germany). A T1-weighted spin echo (SE) sequence was acquired once under resting conditions and another under rigid head motion for all volunteers \citep{Küstner2018}. Similar to the PET-CT dataset, the MR data was scaled to a spacing of $1 \times 1 \times 1 \ \textrm{mm}^{3}$ and 2D axial slices of $256 \times 256$ resolution was extracted from the brain region. Image data were paired in that a motion-free and a motion-corrupted image were acquired and aligned using the header information. The training datasets consisted of 1445 MR images from 7 patients, while evaluation was carried out on a separate dataset of 556 images from 4 patients.

For the final application of this study, the MedGAN framework was utilized for direct denoising of PET imaging data. For this study anonymized datasets were used for the head, torso and abdomen regions from 33 patients using a PET scanner (Biograph mCT, Siemens Healthineers, Germany). The scans have a resolution of $2.8 \times 2.8 \times 2 \ \textrm{mm}^{3}$ and a volume of $256 \times 256 \times 479$. Noisy PET scans, produced by reconstructing PET images from only 25 \% of the original acquisition time, and original PET scans were paired together in a dataset of 11420 training 2D axial slices and 4411 validation images.

\subsection{Experimental setup}

\subsubsection{Analysis of loss functions}

In addition to the conditional adversarial loss, MedGAN incorporates a new combination of non-adversarial losses as part of its framework. Namely, the perceptual, style and content losses. This combination of different loss functions is essential to capture the low frequencies, ensuring global consistency, as well as the high frequency details of the desired target images. The first set of experiments is concerned with studying the impact of individual loss components and showing that MedGAN is more than the sum of its parts. To this end, separate models, each utilizing an individual loss component, were trained and compared with MedGAN for the task of PET-CT translation. For a fair comparison, all trained models utilized identical architectures consisting of a single U-block generator and a $16 \times 16$ patch discriminator network. However, for MedGAN two separate variants were investigated. Specifically, a MedGAN incorporating a CasNet architecture of 6 U-blocks and a MedGAN whose generator consists of only 1 U-block, referred to as MedGAN-1G. This is to illustrate that the performance of MedGAN is not solely due to the increased capacity provided by the CasNet architecture but also the utilized non-adversarial losses.

\subsubsection{Comparison with state-of-the-art techniques}
In the second set of experiments, the performance of MedGAN was investigated on three challenging tasks with no task-specific modifications to the hyperparameters. For this purpose, several translation approaches where re-implemented, trained on the three acquired datasets and compared qualitatively and quantitatively with the MedGAN framework. To ensure a faithful representation of the methods used in the comparative study, a publicly verified implementation of pix2pix was used as basis for the re-implementation of the different approaches \citep{git}.

\begin{figure*}[!ht]
	\begin{minipage}[t]{0.145\linewidth}
		\centering
		\begin{overpic}[width=0.879\textwidth]%
			{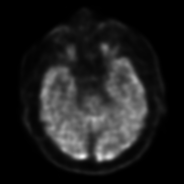}
			\centering
			\put(23,73){Input}
		\end{overpic}	
	\end{minipage}%
	\begin{minipage}[t]{0.705\linewidth}
		\centering
		\begin{overpic}[width=0.1815\textwidth]%
			{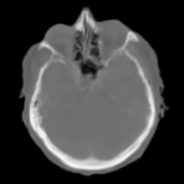}
			\centering
			\put(20,73){cGAN}
		\end{overpic}
		\begin{overpic}[width=0.1815\textwidth]%
			{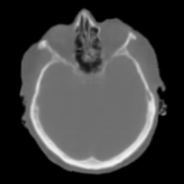}
			\centering
			\put(10.5,73){Perceptual}
		\end{overpic}
		\begin{overpic}[width=0.1815\textwidth]%
			{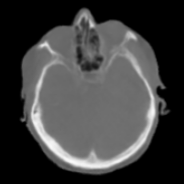}
			\centering
			\put(05,73){Style-content}
		\end{overpic}
		\begin{overpic}[width=0.1815\textwidth]%
			{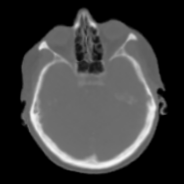}
			\centering
			\put(04,73){MedGAN-1G}
		\end{overpic}
		\begin{overpic}[width=0.1815\textwidth]%
			{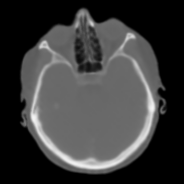}
			\centering
			\put(12,73){MedGAN}
		\end{overpic}
	\end{minipage}
	\begin{minipage}[t]{0.130\linewidth}
		\centering
		\begin{overpic}[width=0.978\textwidth]%
			{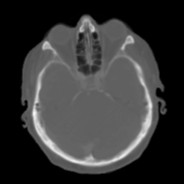}
			\centering
			\put(18,73){Target}
		\end{overpic}		
	\end{minipage}\\
	
	\begin{minipage}[t]{0.145\linewidth}
		\centering
		\begin{overpic}[width=0.879\textwidth]%
			{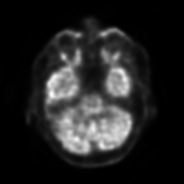}
			\centering
		\end{overpic}	
	\end{minipage}%
	\begin{minipage}[t]{0.705\linewidth}
		\centering
		\begin{overpic}[width=0.1815\textwidth]%
			{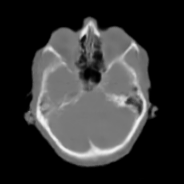}
			\centering
		\end{overpic}
		\begin{overpic}[width=0.1815\textwidth]%
			{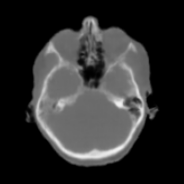}
			\centering
		\end{overpic}
		\begin{overpic}[width=0.1815\textwidth]%
			{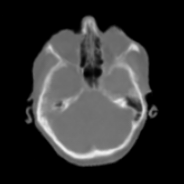}
			\centering
		\end{overpic}
		\begin{overpic}[width=0.1815\textwidth]%
			{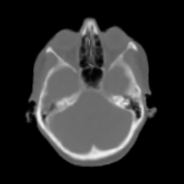}
			\centering
		\end{overpic}
		\begin{overpic}[width=0.1815\textwidth]%
			{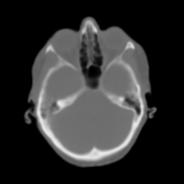}
			\centering
		\end{overpic}
	\end{minipage}
	\begin{minipage}[t]{0.130\linewidth}
		\centering
		\begin{overpic}[width=0.978\textwidth]%
			{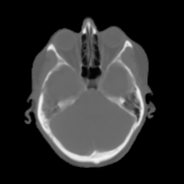}
			\centering
		\end{overpic}		
	\end{minipage}
	\caption{Comparison of the effectiveness of different loss functions used within the MedGAN framework. On the leftmost column, input PET images are given which corresponds to the ground truth CT images given in the rightmost column in two slices. Intermediate columns show synthetically translated CT images as a result of training using different individual loss components.}
	\label{4}
\end{figure*}
First, the cGAN loss was combined with an L1 pixel reconstruction loss into the pix2pix framework \citep{DBLP:journals/corr/IsolaZZE16}. This method was used previously for various medical applications such as MR to CT translation \citep{8310638}, CT denoising \citep{45324} and 2T to 1T MR translation \citep{Yang2018MRIIT}. Moreover, a perceptual adversarial network (PAN) \citep{8358814} was also implemented by incorporating a perceptual loss component similar to that proposed by MedGAN. However, the perceptual loss utilized within the MedGAN framework additionally includes a pixel loss component by calculating the MAE of the raw inputs as well as that of the hidden features extracted by the discriminator. This component was found to be beneficial in maintaining the ensure global consistency of the translated images. Additionally, PAN penalizes the discriminator to preserve the perceptual discrepancy between the hidden features in order to stabilize adversarial training. However, in our experiments, it was found out that such a penalty term often leads to blurry details in the resultant medical images. The Fila-sGAN was developed with a different objective compared to MedGAN. It attempts to transfer the textures of an input style image onto a GAN translated image in order to generate multiple variations of the same underlying structure \citep{Zhao2017SynthesizingFS}. However, it is similar to MedGAN in that it utilizes a pre-trained VGG network to calculate style and content losses in addition to a total variation loss and a L1 pixel reconstruction loss. Therefore, we re-implement Fila-sGAN with the objective of enhanced image translation by replacing the style input images with the original source domain images. The final translation approach used in this comparative study is the ID-CGAN, designed for the denoising of weather artefacts in natural images \citep{DBLP:journals/corr/ZhangSP17}. ID-CGAN incorporates a combination of the adversarial loss, L2 pixel reconstruction loss and the content loss extracted from a pre-trained VGG-19 network. For fair comparison, all methods were trained using the same settings, hyperparameters and architecture (a single U-block generator and patch discriminator) as the MedGAN framework which only differed in utilizing a CasNet generator architecture of 6 U-blocks. 


\subsubsection{Perceptual study and validation}
To judge the fidelity of the translated images, a series of experiments were conducted in which 5 experienced radiologists were presented a series of trials each containing the ground truth target image and the MedGAN output. The main purpose of this study is to investigate how realistic the translated images by MedGAN compared to ground truth medical imagery. However, as a baseline of comparison the same study was repeated for the pix2pix framework. In each of the trails, the images appeared in a randomized order and participants were asked to classify which was the ground truth image as well as rate the quality of each image using a 4-point score, with 4 being the most realistic. Each participant tested one translation application at a time and was presented 60 triads of images from that respective dataset. All images were presented in $256 \times 256$ resolution. 

\begin{table}[!t]
	\caption{\\Quantitative comparison of loss components}
	\centering
	\label{t1}
	\Large
	\bgroup
	\def\arraystretch{1.8}
	\resizebox{\columnwidth}{!}{%
		\begin{tabular}{rcccccc}
			\noalign{\smallskip} \hline \hline \noalign{\smallskip}
			Loss & SSIM & PSNR(dB) & MSE & VIF & UQI & LPIPS \\
			\hline
			cGAN & 0.8960 & 23.65 & 313.2 & 0.3858 & 0.9300 & 0.2592\\
			Peceptual & 0.9071 & 24.20 & 287.0 & 0.4183 & 0.9514 & 0.2628\\
			Style-content & 0.9046 & 24.12 & 282.8 & 0.4105 & 0.9435 & 0.2439\\
			MedGAN-1G & 0.9121 & 24.51 & 271.8 & 0.4348 & \textbf{0.9569} & \textbf{0.2142}
			\\
			MedGAN & \textbf{0.9160} & \textbf{24.62} & \textbf{264.6} & \textbf{0.4464} & 0.9558 & 0.23015
			\\
			\noalign{\smallskip} \hline \noalign{\smallskip}
		\end{tabular}
	}
	\egroup
\end{table}
\subsection{Evaluation metrics}
The performance of the MedGAN framework was evaluated on the above-mentioned datasets both qualitatively and quantitatively. With respect to quantitative experiments, there is no consensus in the scientific community regarding the best evaluation metrics to asses the performance of generative models \citep{DBLP:journals/corr/abs-1802-03446}. Therefore, several image quality metrics were utilized to judge the quality of the translated medical images such as Structural Similarity Index (SSIM) \citep{1284395}, Peak Signal to Noise Ratio (PSNR), Mean Squared Error (MSE), Visual Information Fidelity (VIF) \citep{1576816} and Universal Quality Index (UQI)\citep{995823}. Nevertheless, recent studies pointed out that these metrics could not be counted upon solely as reference for human judgement of image quality. Hence, the recent metric titled Learned Perceptual Image Patch Similarity (LPIPS) was utilized, which was reported to outperform previous metrics as a perceptual measure of quality \citep{DBLP:journals/corr/abs-1801-03924}. For the qualitative comparisons, we present the input, transformed and ground-truth target images. 
\section{Results}

\subsection{Analysis of loss functions}

\begin{figure*}[!t]
	\begin{minipage}[t]{1.0\linewidth}
		\centering
		\begin{minipage}[t]{0.145\linewidth}
			\centering
			\begin{overpic}[width=0.879\textwidth]%
				{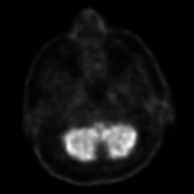}
				\centering
				\put(23,73){Input}
			\end{overpic}	
		\end{minipage}%
		\begin{minipage}[t]{0.705\linewidth}
			\centering
			\begin{overpic}[width=0.1815\textwidth]%
				{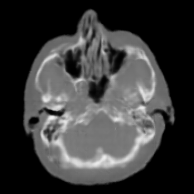}
				\centering
				\put(20,73){pix2pix}
			\end{overpic}
			\begin{overpic}[width=0.1815\textwidth]%
				{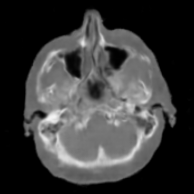}
				\centering
				\put(22,73){PAN}
			\end{overpic}
			\begin{overpic}[width=0.1815\textwidth]%
				{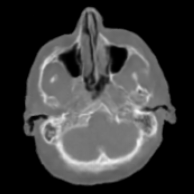}
				\centering
				\put(10,73){ID-CGAN}
			\end{overpic}
			\begin{overpic}[width=0.1815\textwidth]%
				{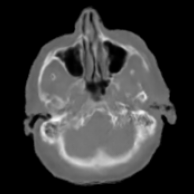}
				\centering
				\put(10,73){Fila-sGAN}
			\end{overpic}
			\begin{overpic}[width=0.1815\textwidth]%
				{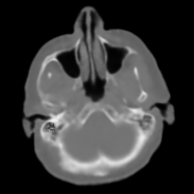}
				\centering
				\put(14,73){MedGAN}
			\end{overpic}
		\end{minipage}
		\begin{minipage}[t]{0.130\linewidth}
			\centering
			\begin{overpic}[width=0.978\textwidth]%
				{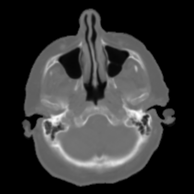}
				\centering
				\put(18,73){Target}
			\end{overpic}		
		\end{minipage}\\
		\vspace{4mm}
		\begin{minipage}[t]{0.145\linewidth}
			\centering
			\begin{overpic}[width=0.879\textwidth]%
				{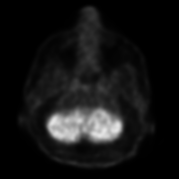}
				\centering
			\end{overpic}	
		\end{minipage}%
		\begin{minipage}[t]{0.705\linewidth}
			\centering
			\begin{overpic}[width=0.1815\textwidth]%
				{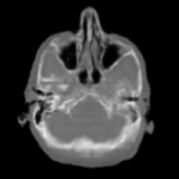}
				\centering
			\end{overpic}
			\begin{overpic}[width=0.1815\textwidth]%
				{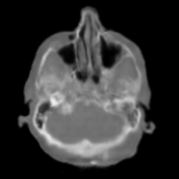}
				\centering
			\end{overpic}
			\begin{overpic}[width=0.1815\textwidth]%
				{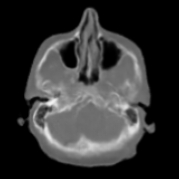}
				\centering
			\end{overpic}
			\begin{overpic}[width=0.1815\textwidth]%
				{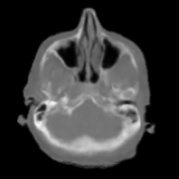}
				\centering
			\end{overpic}
			\begin{overpic}[width=0.1815\textwidth]%
				{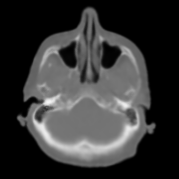}
				\centering
			\end{overpic}
		\end{minipage}
		\begin{minipage}[t]{0.130\linewidth}
			\centering
			\begin{overpic}[width=0.978\textwidth]%
				{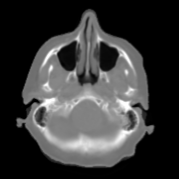}
				\centering
			\end{overpic}		
		\end{minipage}
		(a) PET-CT translation
	\end{minipage}\\
	
	\vspace{9mm}

	\begin{minipage}[t]{1.0\linewidth}
		\centering
		\begin{minipage}[t]{0.145\linewidth}
			\centering
			\begin{overpic}[width=0.879\textwidth]%
				{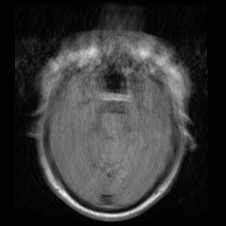}
				\centering
				\put(23,73){Input}
			\end{overpic}	
		\end{minipage}%
		\begin{minipage}[t]{0.705\linewidth}
			\centering
			\begin{overpic}[width=0.1815\textwidth]%
				{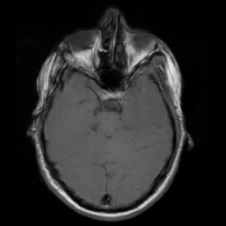}
				\centering
				\put(20,73){pix2pix}
			\end{overpic}
			\begin{overpic}[width=0.1815\textwidth]%
				{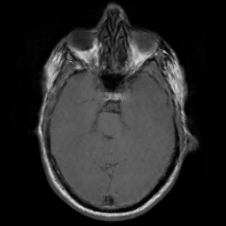}
				\centering
				\put(22,73){PAN}
			\end{overpic}
			\begin{overpic}[width=0.1815\textwidth]%
				{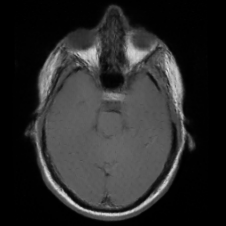}
				\centering
				\put(10,73){ID-CGAN}
			\end{overpic}
			\begin{overpic}[width=0.1815\textwidth]%
				{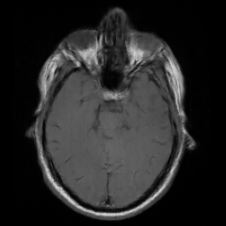}
				\centering
				\put(10,73){Fila-sGAN}
			\end{overpic}
			\begin{overpic}[width=0.1815\textwidth]%
				{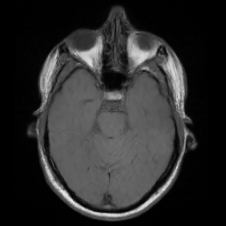}
				\centering
				\put(14,73){MedGAN}
			\end{overpic}
		\end{minipage}
		\begin{minipage}[t]{0.130\linewidth}
			\centering
			\begin{overpic}[width=0.978\textwidth]%
				{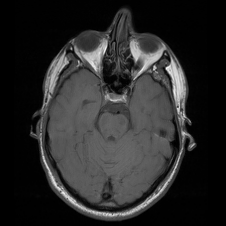}
				\centering
				\put(18,73){Target}
			\end{overpic}		
		\end{minipage}\\
		\vspace{4mm}
		\begin{minipage}[t]{0.145\linewidth}
			\centering
			\begin{overpic}[width=0.879\textwidth]%
				{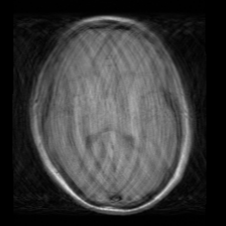}
				\centering
			\end{overpic}	
		\end{minipage}%
		\begin{minipage}[t]{0.705\linewidth}
			\centering
			\begin{overpic}[width=0.1815\textwidth]%
				{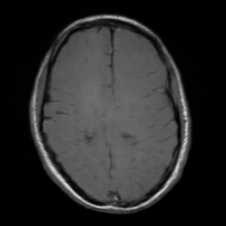}
				\centering
			\end{overpic}
			\begin{overpic}[width=0.1815\textwidth]%
				{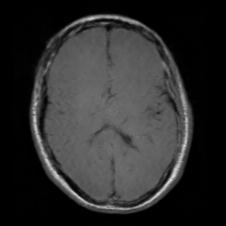}
				\centering
			\end{overpic}
			\begin{overpic}[width=0.1815\textwidth]%
				{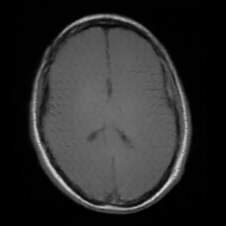}
				\centering
			\end{overpic}
			\begin{overpic}[width=0.1815\textwidth]%
				{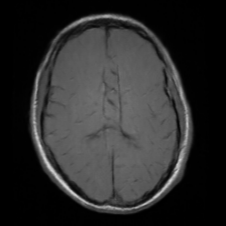}
				\centering
			\end{overpic}
			\begin{overpic}[width=0.1815\textwidth]%
				{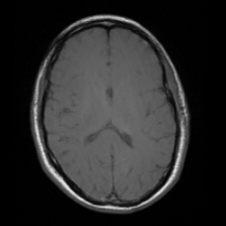}
				\centering
			\end{overpic}
		\end{minipage}
		\begin{minipage}[t]{0.130\linewidth}
			\centering
			\begin{overpic}[width=0.978\textwidth]%
				{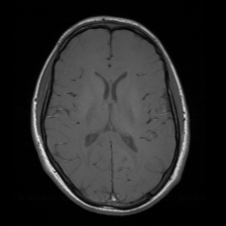}
				\centering
			\end{overpic}		
		\end{minipage}
		(b) MR motion correction
	\end{minipage}
	
	\vspace{9mm}
	
	\begin{minipage}[t]{1.0\linewidth}
		\centering
		\begin{minipage}[t]{0.145\linewidth}
			\centering
			\begin{overpic}[width=0.879\textwidth]%
				{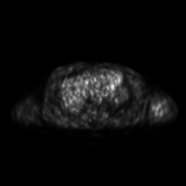}
				\centering
				\put(23,73){Input}
			\end{overpic}	
		\end{minipage}%
		\begin{minipage}[t]{0.705\linewidth}
			\centering
			\begin{overpic}[width=0.1815\textwidth]%
				{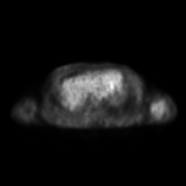}
				\centering
				\put(20,73){pix2pix}
			\end{overpic}
			\begin{overpic}[width=0.1815\textwidth]%
				{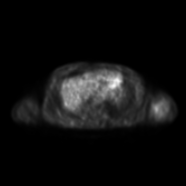}
				\centering
				\put(22,73){PAN}
			\end{overpic}
			\begin{overpic}[width=0.1815\textwidth]%
				{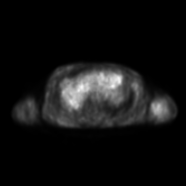}
				\centering
				\put(10,73){ID-CGAN}
			\end{overpic}
			\begin{overpic}[width=0.1815\textwidth]%
				{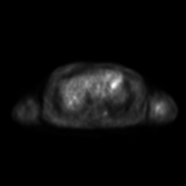}
				\centering
				\put(10,73){Fila-sGAN}
			\end{overpic}
			\begin{overpic}[width=0.1815\textwidth]%
				{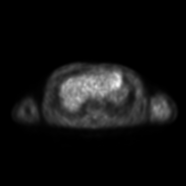}
				\centering
				\put(14,73){MedGAN}
			\end{overpic}
		\end{minipage}
		\begin{minipage}[t]{0.130\linewidth}
			\centering
			\begin{overpic}[width=0.978\textwidth]%
				{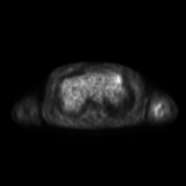}
				\centering
				\put(18,73){Target}
			\end{overpic}		
		\end{minipage}\\
		\vspace{4mm}
		\begin{minipage}[t]{0.145\linewidth}
			\centering
			\begin{overpic}[width=0.879\textwidth]%
				{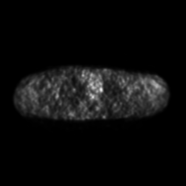}
				\centering
			\end{overpic}	
		\end{minipage}%
		\begin{minipage}[t]{0.705\linewidth}
			\centering
			\begin{overpic}[width=0.1815\textwidth]%
				{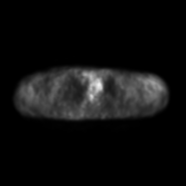}
				\centering
			\end{overpic}
			\begin{overpic}[width=0.1815\textwidth]%
				{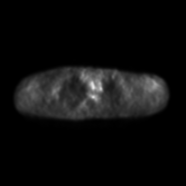}
				\centering
			\end{overpic}
			\begin{overpic}[width=0.1815\textwidth]%
				{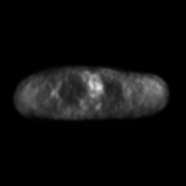}
				\centering
			\end{overpic}
			\begin{overpic}[width=0.1815\textwidth]%
				{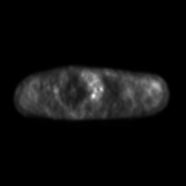}
				\centering
			\end{overpic}
			\begin{overpic}[width=0.1815\textwidth]%
				{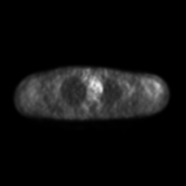}
				\centering
			\end{overpic}
		\end{minipage}
		\begin{minipage}[t]{0.130\linewidth}
			\centering
			\begin{overpic}[width=0.978\textwidth]%
				{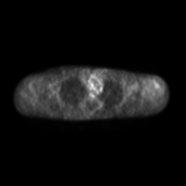}
				\centering
			\end{overpic}		
		\end{minipage}
		(c) PET denoising
	\end{minipage}
	\caption{Comparison between MedGAN and different image translation approaches for the proposed medical image translation tasks. Two respective image slices are shown for each task.}
	\label{5}
\end{figure*}

\begin{table*}[htb]
	\caption{\\Quantitative comparison between MedGAN and other translation frameworks}
	\centering
	\label{t2}
	\Huge
	\bgroup
	\def\arraystretch{1.7}
	\begin{adjustbox}{width=1.0\textwidth}
		\begin{tabular}{r|cccccc|cccccc|cccccc}
			\noalign{\smallskip} \hline \hline \noalign{\smallskip}
			\multirow{2}{*}{Method} & \multicolumn{6}{c}{(a) PET-CT translation} & %
			\multicolumn{6}{c}{(b) MR motion correction} & \multicolumn{6}{c}{(c) PET denoising}\\
			& SSIM & PSNR(dB) & MSE & VIF & UQI & LPIPS & SSIM & PSNR(dB) & MSE & VIF & UQI & LPIPS & SSIM & PSNR(dB) & MSE & VIF & UQI & LPIPS\\
			\hline
			pix2pix & 0.9017 & 23.93 & 299.2 & 0.4024 & 0.9519 & 0.2537 & 
			0.8138 & 23.79 & 335.2 & 0.3464 & 0.5220 & 0.2885 
			& 0.9707 & 34.89 & 37.20 & 0.6068 & 0.9440 & 0.0379\\
			
			PAN & 0.9027 & 24.10 & 292.2 & 0.4084 & 0.9190 & 0.2582 
			& 0.8116 & 23.91 & 311.4 & 0.3548 & 0.5399 & 0.2797 
			& 0.9713 & 34.97 & 38.76 & 0.6068 & 0.9431 & 0.0348\\
			
			ID-CGAN & 0.9039 & 24.13 & 288.6 & 0.4059 & 0.9389 & 0.2423 
			& 0.8214 & \textbf{24.26} & 289.8 & 0.3685 & 0.5855 & 0.2747 
			& 0.9699 & 34.28 & 39.45 & 0.6023 & 0.9435 & 0.0346\\
			
			
			Fila-sGAN & 0.9039 & 24.08 & 289.6 & 0.4146 & 0.9054 & 0.2320 
			& 0.8114 & 23.91 & 318.7 & 0.3431 & 0.4957 & 0.2570
			& 0.9726 & 35.05 & 35.80 & \textbf{0.6279} & \textbf{0.9472} & 0.0328\\
			
			MedGAN & \textbf{0.9160} & \textbf{24.62} & \textbf{264.6} & \textbf{0.4464} & \textbf{0.9558} & \textbf{0.2302} & \textbf{0.8363} & 24.18 & \textbf{289.9} & \textbf{0.3735} & \textbf{0.6037} & \textbf{0.2391}
			& \textbf{0.9729} & \textbf{35.23} & \textbf{33.54} & 0.6168 & 0.9443 & \textbf{0.0292}\\
			
			\noalign{\smallskip} \hline \noalign{\smallskip}
		\end{tabular}
	\end{adjustbox}
	\egroup
\end{table*}

The results of utilizing individual loss functions in comparison to MedGAN are presented in \hyperref[4]{Fig.~\ref*{4}} and \hyperref[t1]{Table~I} respectively. From a qualitative point of view, it was found out that the traditional adversarial loss $\mathcal{L}_{\small\textrm{cGAN}}$ leads to the worst results (\hyperref[4]{Fig.~\ref*{4}}). This is also reflected in the quantitative scores (\hyperref[t1]{Table~I}) where cGAN achieves the worst numerical scores across the chosen metrics. On the other hand, the perceptual loss improves the results by enhancing the details of the resultant bone structures. It also refines the global consistency due to the pixel-wise component of the loss. However, when compared to the ground truth target images, it is observed that the translated CT images have a reduced level of details. Combining the generative framework with a pre-trained feature extractor (VGG-19) for the calculation of style and content losses further improves the qualitative results. This is reflected by the transformed images having sharper details and more fine-tuned structures due to matching the target's textural and global content. The MedGAN-1G framework results in an increased sharpness of the translated images as well as a notable improvement of the quantitative metrics compared to the individual loss components. Yet, incorporating the CasNet generator architecture further enhances the translated output iamges with more refined bone structures and details. As shown in Table I, this is reflected by a significant reduction in the MSE as well as increases in the SSIM, PSNR and VIF compared to MedGAN-1G with a reactively small difference in the UQI and LPIPS scores. 
\subsection{Comparison with state-of-the-art techniques}

For the second set of experiments, the performance of MedGAN was compared against several state-of-the-art translation frameworks including pix2pix, PAN, ID-CGAN and Fila-sGAN. The results are given in \hyperref[5]{Fig.~\ref*{5}} and \hyperref[t2]{Table~II} for the qualitative and quantitative comparisons, respectively. Pix2pix produces the worst results with PAN only slightly outperforming it. 
For MR motion correction, pix2pix and PAN succeeded in producing globally consistent MR images, albeit with blurry details. However, for PET to CT translation the output images lacked sharpness and homogeneity, including realistic bone structures. This was also reflected quantitatively with these methods achieving the worst scores in \hyperref[t2]{Table~II}. ID-CGAN outperformed the previous methods in PET to CT translation with the resultant images having a more consistent global structure. However, ID-CGAN did not perform as strongly on the other datasets. For example, ID-CGAN resulted in significant tilting artefacts as well as blurred output details in MR motion correction. Similarly, Fila-sGAN produced inconsistent results on the different datasets. While it produced positive results in PET to CT translation, Fila resulted in blurred denoised PET images and unrealistic textures in the motion corrected MR images. The MedGAN framework outperformed the other approaches on the three different translation tasks. It produces sharper and more homogeneous outputs from the visual perspective. The performance of MedGAN was also reflected quantitatively in \hyperref[t2]{Table~II}. It resulted in the best score for the different tasks across the large majority of the chosen metrics.

\subsection{Perceptual study and validation}
The results of the perceptual study conducted by radiologists on the three utilized datasets are presented in \hyperref[t3]{Table~III}. The final column of this table states the percentage of images classified by radiologists as real out of the triad of presented images. In the PET to CT translation, 25.3 \% of the synthetically generated CT images by the MedGAN framework managed to convince radiologists into thinking they are ground truth images from a real CT scanner. In MR motion correction and PET denoising, the percentage of MedGAN images classified as real was 6.7 \% and 14.3 \% respectively. Additionally, radiologists rated the output of the MedGAN framework highly with a mean score of 3.22 in comparison to 1.70 achieved by pix2pix and 3.81 by the ground truth images. The performance of MedGAN was also reflected in the remaining two applications, where MedGAN achieved a mean score of 2.81 in comparison to 1.98, and a score of 3.02 in comparison to 1.73 by pix2pix in MR motion correction and PET denoising, respectively.
\begin{table}[t]
	\caption{\\Results of perceptual study}
	\centering
	\label{t3}
	\setlength\arrayrulewidth{0.05pt}
	\tiny
	\bgroup
	\def\arraystretch{1.1}
	\resizebox{\columnwidth}{!}{%
		\begin{adjustbox}{width=0.1\textwidth}
			\begin{tabular}{r|ccc}
				\hline
				\multirow{2}{*}{Method} & \multicolumn{3}{c}{(a) PET-CT translation}\\
				& mean & SD & real \% \\
				\hline
				pix2pix & 1.70 & 0.531 & 0.00\\			
				MedGAN & 3.22 & 0.590 & 25.3\\
				Ground truth & 3.81 & 0.394 & 74.7\\
				\hline \hline
				\multirow{2}{*}{Method} & \multicolumn{3}{c}{(b) MR motion correction}\\
				& mean & SD & real \% \\ \hline
				pix2pix & 1.98 & 0.514 & 0.00\\			
				MedGAN & 2.81 & 0.573 & 6.70\\
				Ground truth & 3.87 & 0.337 & 93.3\\
				\hline \hline
				\multirow{2}{*}{Method} & \multicolumn{3}{c}{(c) PET denoising}\\
				& mean & SD & real \% \\ \hline
				pix2pix & 1.73 & 0.511 & 0.00\\			
				MedGAN & 3.02 & 0.542 & 14.3\\
				Ground truth & 3.70 & 0.461 & 85.7\\			
				\hline \noalign{\smallskip}
			\end{tabular}
		\end{adjustbox}
	}
	\egroup
\end{table}

\section{Discussion}

In this work, MedGAN was presented as an end-to-end framework for medical image translation tasks. MedGAN incorporates a new combination of non-adversarial losses, namely the perceptual and style-content losses, on top of an adversarial framework to capture the high and low frequency components of the target images. The proposed framework utilizes the CasNet architecture, a generator network which progressively refines the translated image via encoder-decoder pairs in an end-to-end manner. This leads to homogeneous and realistic global structures as well as fine-tuned textures and details.

An analysis performed on the task of PET to CT translation, presented in \hyperref[4]{Fig.~\ref*{4}} and \hyperref[t1]{Table~I}, illustrated that MedGAN surpasses the performance of its individual loss components. The cGAN framework results in the worst performance both qualitatively and quantitatively. Specifically, the resulting CT images of this method have a largely non-homogeneous global structure compared to the desired ground truth images. A good example would be examining the bone structures of the nose region in the resultant images. Comparatively, the utilization of the perceptual loss and the style-content losses resulted in an overall improved performance. However, it was observed that the style-content losses have a more significant impact upon the quality of the resultant images. Nevertheless, this impact was not reflected in the quantitative results in \hyperref[t1]{Table~I} where the perceptual loss excels in comparison. This may be attributed to the reported fact that the quantitative scores may not always reflect the perceptual quality of human judgement \citep{DBLP:journals/corr/abs-1801-03924}. The proposed MedGAN framework combines the above mentioned benefits of individual loss components, as it jointly ensures global homogeneity of the resultant images and enhances the level of output details. This improvement is not only the result of the increased capacity provided by the CasNet architecture. MedGAN-1G, with a single U-block generator, also surpasses qualitatively and quantitatively the results of models utilizing individual loss components and identical architectures. Further analysis of the performance of the CasNet architecture is presented in Appendix A.

MedGAN was directly applied with no application-specific modifications on three translation tasks in medical imaging: PET to CT translation, correction of motion artefacts in MR images and PET denoising. For each of these tasks, the performance of MedGAN was compared against other image translation approaches. In the task of PET to CT translation, MedGAN produced realistic and homogeneous bone structures in the resultant CT images that closely matched the ground truth CT images and surpasses visually those produced by ID-CGAN and Fila-sGAN. In the task of MR motion correction, the resultant synthetic MR images are artefact-free with realistic textures and fine-structures. Finally, in the task of PET image denoising, MedGAN produced sharp denoised images as opposed to the blurred results by the other methods. Qualitative comparisons are highly subjective and can not be relied solely upon. Nevertheless, quantitative assessments, given in \hyperref[t2]{Table~II}, also reflect the above conclusions with MedGAN outperforming the other approaches cross the majority of the utilized metrics for the different translation tasks. Additional results are presented in Appendix B.

Also, a perceptual study conducted by 5 radiologists illustrated the fidelity of the translated images by MedGAN. The quality of the output images by MedGAN was rated between $2.8-3.2$ out of a scale of 4. For reference, the ground truth images were rated between $3.7-3.8$ and images from the pix2pix framework were rated between $1.7-2.0$. Furthermore, a subset of $6-25$ \% of the images in the study convinced the radiologists into thinking they are more realistic the ground truth images.

This work is not free from limitations, with further improvements essential for practical applicability of MedGAN in medical post-processing tasks. Translation of 2D slices is substantially more computationally efficient than 3D. Thus, operating in 2D was advantageous for the purpose of this work as it enabled efficient experimentations on different loss functions, architectures and regularization techniques. However, volumetric information in 3D data is essential for the majority of medical tasks. Therefore, in the future, MedGAN will be appropriately adapted to operate on 3D medical volumes. Moreover, medical acquisitions typically result in multi-channel volumes. In this work, only single-channel inputs were considered for computational reasons. However, this is detrimental for tasks such as MR motion correction where the phase information is important for the accurate correction of motion artefacts. In the future, we aim to overcome this disadvantage by expanding the MedGAN framework to accommodate multi-channel inputs. Finally, the main purpose of the MedGAN framework is enhancing technical post-processing tasks that require globally consistent image properties. At this stage, it is not suitable for diagnostic applications. Future research efforts will be directed towards investigating the possibility of reaching diagnostic quality.

\section{Conclusion} 

MedGAN is a new end-to-end framework for medical image translation tasks. It combines the conditional adversarial framework with a new combination of non-adversarial losses and a CasNET generator architecture to enhance the global consistency and high frequency details of results. MedGAN was applied with no task-specific modifications on three challenging tasks in medical imaging: PET-CT translation, MR motion correction and PET denoising. The proposed framework outperformed other similar translation approaches quantitatively and qualitatively across the different proposed tasks. Finally, the subjective performance and fidelity of MedGAN's results were positively attested by 5 experienced radiologists.

Future efforts will be directed for the extension of MedGAN to accommodate 3D multi-channel volumes. Additionally, the performance of MedGAN in technical post-processing tasks will be investigated. For instance, the utilization of synthetically translated CT images for the attenuation correction of PET volumes. Also, we plan to explore the applicability of utilizing retrospectively corrected MR images in a large cohort for segmentation and organ volume calculation.

\ifCLASSOPTIONcaptionsoff
  \newpage
\fi



\bibliographystyle{plainnat}
%

%


\balance
{\footnotesize
	\bibliography{refs}}

\appendices

\clearpage
%
\nobalance
\makeatletter
\renewcommand\section{\@startsection{section}{1}{\z@}{1.5ex plus 1.5ex minus 0.5ex}%
	{0.7ex plus 1ex minus 0ex}{\vskip 1.3em\Huge\centering}}
\makeatother

\appendices
\twocolumn[\section{\textbf{Analysis of Generator Architecture}\vspace{1.3\baselineskip}}\label{app:t1}]

\begin{center}
	\begin{minipage}[h]{1\linewidth}
		\centering
		\includegraphics[width=1\textwidth]{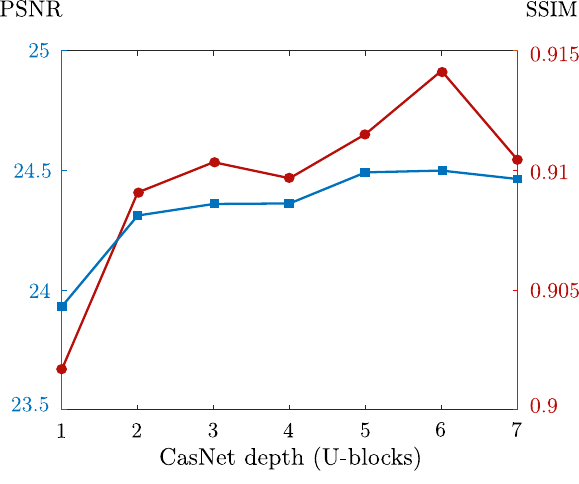}		
		
		(a) PSNR and SSIM scores
	\end{minipage}%
	\begin{minipage}[h]{1\linewidth}
		\centering
		\includegraphics[width=1.015\textwidth]{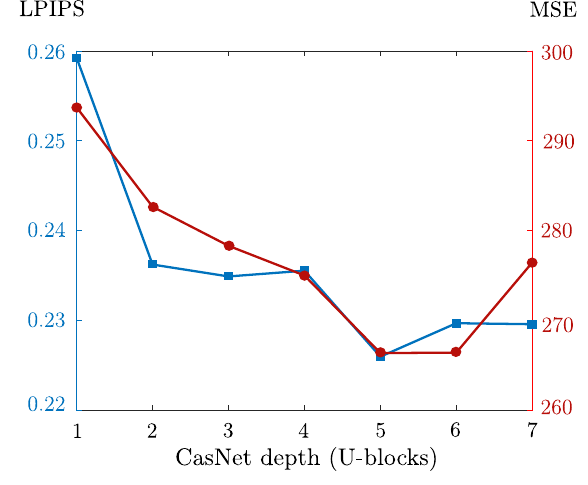}
		
		(b) MSE and LPIPS scores
	\end{minipage}
	\captionof{figure}{Quantitative performance of a pix2p\rlap{ix network using a CasNet with a varying number of U-blocks.}}
	\label{6}
\end{center}

%
%
%

\begin{center}
	\vspace{10mm}
	\rlap{\begin{minipage}[t]{0.29\linewidth}
		\centering
		\begin{overpic}[width=0.879\textwidth]%
			{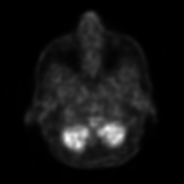}
			\centering
			\put(23,73){Input}
		\end{overpic}	
	\end{minipage}%
	\begin{minipage}[t]{1.41\linewidth}
		\centering
		\begin{overpic}[width=0.1815\textwidth]%
			{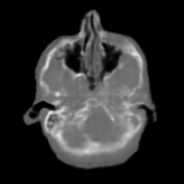}
			\centering
			\put(10,73){pix2pix-1G}
		\end{overpic}
		\begin{overpic}[width=0.1815\textwidth]%
			{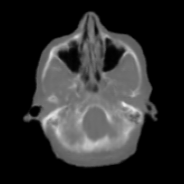}
			\centering
			\put(10,73){pix2pix-3G}
		\end{overpic}
		\begin{overpic}[width=0.1815\textwidth]%
			{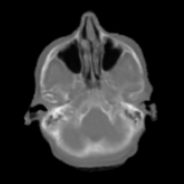}
			\centering
			\put(10,73){pix2pix-5G}
		\end{overpic}
		\begin{overpic}[width=0.1815\textwidth]%
			{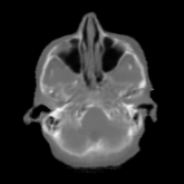}
			\centering
			\put(10,73){pix2pix-6G}
		\end{overpic}
		\begin{overpic}[width=0.1815\textwidth]%
			{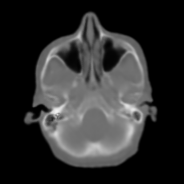}
			\centering
			\put(14,73){MedGAN}
		\end{overpic}
	\end{minipage} 
	\begin{minipage}[t]{0.26\linewidth}
		\centering
		\begin{overpic}[width=0.978\textwidth]%
			{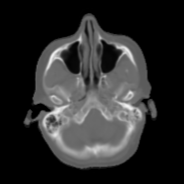}
			\centering
			\put(18,73){Target}
		\end{overpic}		
	\end{minipage}}
	\vspace{4mm}
	\rlap{\begin{minipage}[t]{0.29\linewidth}
		\centering
		\begin{overpic}[width=0.879\textwidth]%
			{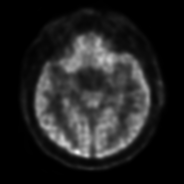}
			\centering
		\end{overpic}	
	\end{minipage}%
	\begin{minipage}[t]{1.41\linewidth}
		\centering
		\begin{overpic}[width=0.1815\textwidth]%
			{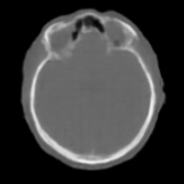}
			\centering
		\end{overpic}
		\begin{overpic}[width=0.1815\textwidth]%
			{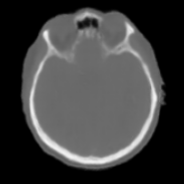}
			\centering
		\end{overpic}
		\begin{overpic}[width=0.1815\textwidth]%
			{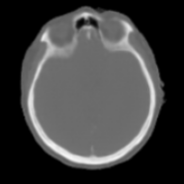}
			\centering
		\end{overpic}
		\begin{overpic}[width=0.1815\textwidth]%
			{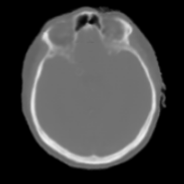}
			\centering
		\end{overpic}
		\begin{overpic}[width=0.1815\textwidth]%
			{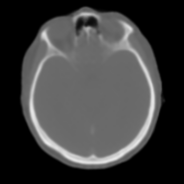}
			\centering
		\end{overpic}
	\end{minipage}
	\begin{minipage}[t]{0.26\linewidth}
		\centering
		\begin{overpic}[width=0.978\textwidth]%
			{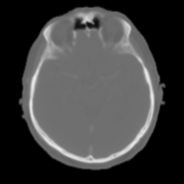}
			\centering
		\end{overpic}		
	\end{minipage}}
	\vspace{4mm}
		\rlap{\begin{minipage}[t]{0.29\linewidth}
			\centering
			\begin{overpic}[width=0.879\textwidth]%
				{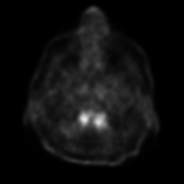}
				\centering
			\end{overpic}	
		\end{minipage}%
		\begin{minipage}[t]{1.41\linewidth}
			\centering
			\begin{overpic}[width=0.1815\textwidth]%
				{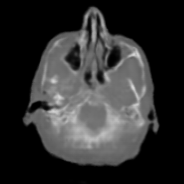}
				\centering
			\end{overpic}
			\begin{overpic}[width=0.1815\textwidth]%
				{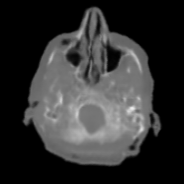}
				\centering
			\end{overpic}
			\begin{overpic}[width=0.1815\textwidth]%
				{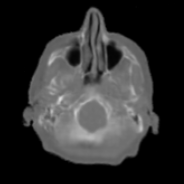}
				\centering
			\end{overpic}
			\begin{overpic}[width=0.1815\textwidth]%
				{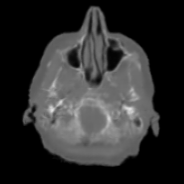}
				\centering
			\end{overpic}
			\begin{overpic}[width=0.1815\textwidth]%
				{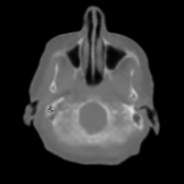}
				\centering
			\end{overpic}
		\end{minipage}
		\begin{minipage}[t]{0.26\linewidth}
			\centering
			\begin{overpic}[width=0.978\textwidth]%
				{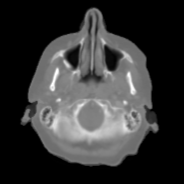}
				\centering
			\end{overpic}		
		\end{minipage}}
	\captionof{figure}{Comparison of pix2pix with a CasNet of 1, 3, 5 an\rlap{d 6 U-blocks, respectively, versus the MedGAN framework.}}
	\label{7}
\end{center}




To investigate the performance of the proposed CasNet architecture, we implemented pix2pix models utilizing CasNet with different block depth, from 1 to 7 U-blocks. Quantitative performance is presented in \hyperref[6]{Fig.~\ref*{6}}. It can be seen that as the CasNet utilizes greater capacity through a larger concatenation of U-blocks, quantitative performance increases significantly up until the 6th U-block. Beyond this point, performance either saturates,\\ \\ \\ \\ \\ \\ \\ \\ \\ \\ \\ \\ \\ \\ \\ \\ \\ \\ \\ \\   \\ \\ \\ \\ \\ \\ \\ \\ \\ \\ \\ \\ \\ \\ \\ \\ \\ \\ \\ \\ \\ \\ \\ \\  \\ \\ \;  e.g. PSNR and MSE, or starts to degrade, in the case of SSIM and LPIPS scores. Further investigations are required to determine the optimum depth of CasNet. \hyperref[7]{Fig.~\ref*{7}} illustrates the effect of CasNet on the translated images by pix2pix of different CasNet depth in comparison to MedGAN (6 U-blocks). It can be seen that as the number of U-blocks increases visual quality of translated images is significantly improved.

\twocolumn[\section{\textbf{Additional Results}\vspace{1.3\baselineskip}}\label{app:t1}]

\begin{center}
	\begin{minipage}[t]{1.0\linewidth}
		\vspace{5mm}
		\centering
		\rlap{\begin{minipage}[t]{0.29\linewidth}
			\centering
			\begin{overpic}[width=0.879\textwidth]%
				{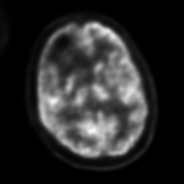}
				\centering
				\put(23,73){Input}
			\end{overpic}	
		\end{minipage}%
		\begin{minipage}[t]{1.41\linewidth}
			\centering
			\begin{overpic}[width=0.1815\textwidth]%
				{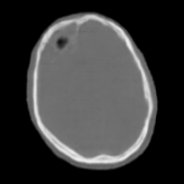}
				\centering
				\put(20,73){pix2pix}
			\end{overpic}
			\begin{overpic}[width=0.1815\textwidth]%
				{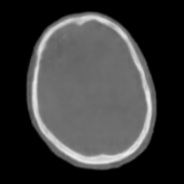}
				\centering
				\put(22,73){PAN}
			\end{overpic}
			\begin{overpic}[width=0.1815\textwidth]%
				{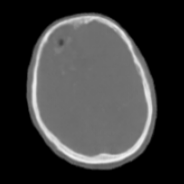}
				\centering
				\put(10,73){ID-CGAN}
			\end{overpic}
			\begin{overpic}[width=0.1815\textwidth]%
				{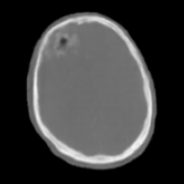}
				\centering
				\put(10,73){Fila-sGAN}
			\end{overpic}
			\begin{overpic}[width=0.1815\textwidth]%
				{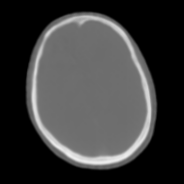}
				\centering
				\put(14,73){MedGAN}
			\end{overpic}
		\end{minipage}
		\begin{minipage}[t]{0.26\linewidth}
			\centering
			\begin{overpic}[width=0.978\textwidth]%
				{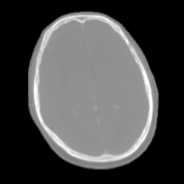}
				\centering
				\put(18,73){Target}
			\end{overpic}		
		\end{minipage}}
		\vspace{4mm}
		\rlap{\begin{minipage}[t]{0.29\linewidth}
			\centering
			\begin{overpic}[width=0.879\textwidth]%
				{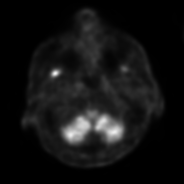}
				\centering
			\end{overpic}	
		\end{minipage}%
		\begin{minipage}[t]{1.41\linewidth}
			\centering
			\begin{overpic}[width=0.1815\textwidth]%
				{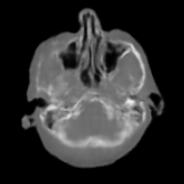}
				\centering
			\end{overpic}
			\begin{overpic}[width=0.1815\textwidth]%
				{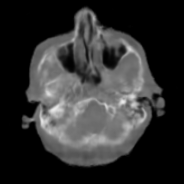}
				\centering
			\end{overpic}
			\begin{overpic}[width=0.1815\textwidth]%
				{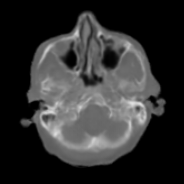}
				\centering
			\end{overpic}
			\begin{overpic}[width=0.1815\textwidth]%
				{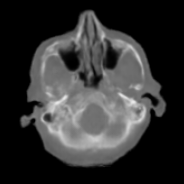}
				\centering
			\end{overpic}
			\begin{overpic}[width=0.1815\textwidth]%
				{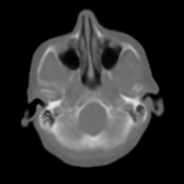}
				\centering
			\end{overpic}
		\end{minipage}
		\begin{minipage}[t]{0.26\linewidth}
			\centering
			\begin{overpic}[width=0.978\textwidth]%
				{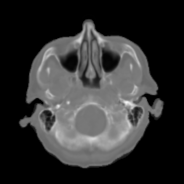}
				\centering
			\end{overpic}		
		\end{minipage}}
	
		\hspace{0.62\linewidth}\rlap{(a) PET-CT translation} 
	\end{minipage}	

	\vspace{5mm}
	
	\begin{minipage}[t]{1.0\linewidth}
		\vspace{5mm}
		\centering
		\rlap{\begin{minipage}[t]{0.29\linewidth}
				\centering
				\begin{overpic}[width=0.879\textwidth]%
					{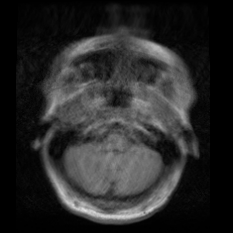}
					\centering
					\put(23,73){Input}
				\end{overpic}	
			\end{minipage}%
			\begin{minipage}[t]{1.41\linewidth}
				\centering
				\begin{overpic}[width=0.1815\textwidth]%
					{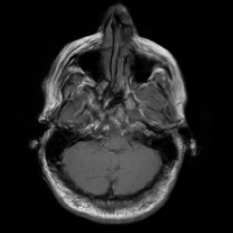}
					\centering
					\put(20,73){pix2pix}
				\end{overpic}
				\begin{overpic}[width=0.1815\textwidth]%
					{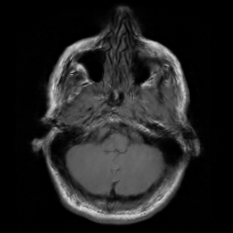}
					\centering
					\put(22,73){PAN}
				\end{overpic}
				\begin{overpic}[width=0.1815\textwidth]%
					{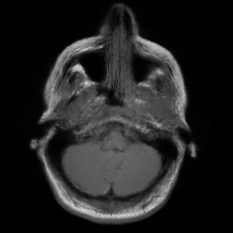}
					\centering
					\put(10,73){ID-CGAN}
				\end{overpic}
				\begin{overpic}[width=0.1815\textwidth]%
					{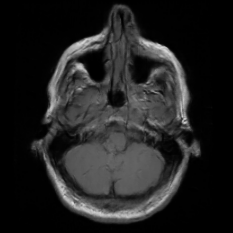}
					\centering
					\put(10,73){Fila-sGAN}
				\end{overpic}
				\begin{overpic}[width=0.1815\textwidth]%
					{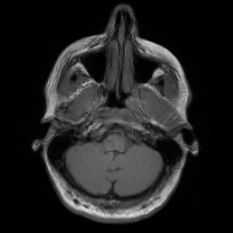}
					\centering
					\put(14,73){MedGAN}
				\end{overpic}
			\end{minipage}
			\begin{minipage}[t]{0.26\linewidth}
				\centering
				\begin{overpic}[width=0.978\textwidth]%
					{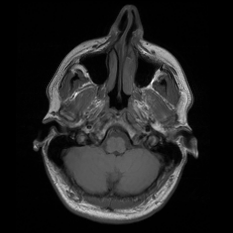}
					\centering
					\put(18,73){Target}
				\end{overpic}		
		\end{minipage}}
		\vspace{4mm}
		\rlap{\begin{minipage}[t]{0.29\linewidth}
				\centering
				\begin{overpic}[width=0.879\textwidth]%
					{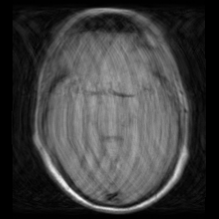}
					\centering
				\end{overpic}	
			\end{minipage}%
			\begin{minipage}[t]{1.41\linewidth}
				\centering
				\begin{overpic}[width=0.1815\textwidth]%
					{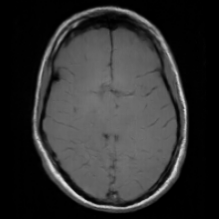}
					\centering
				\end{overpic}
				\begin{overpic}[width=0.1815\textwidth]%
					{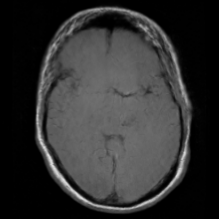}
					\centering
				\end{overpic}
				\begin{overpic}[width=0.1815\textwidth]%
					{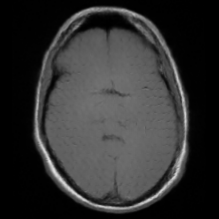}
					\centering
				\end{overpic}
				\begin{overpic}[width=0.1815\textwidth]%
					{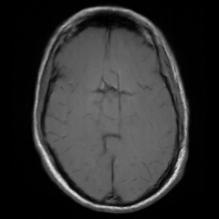}
					\centering
				\end{overpic}
				\begin{overpic}[width=0.1815\textwidth]%
					{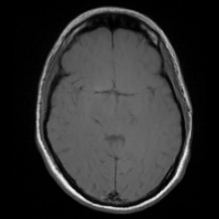}
					\centering
				\end{overpic}
			\end{minipage}
			\begin{minipage}[t]{0.26\linewidth}
				\centering
				\begin{overpic}[width=0.978\textwidth]%
					{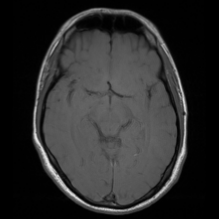}
					\centering
				\end{overpic}		
		\end{minipage}}
		
		\hspace{0.62\linewidth}\rlap{(b) MR motion correction } 
	\end{minipage}	
	
	\vspace{5mm}
	
	\begin{minipage}[t]{1.0\linewidth}
		\vspace{5mm}
		\centering
		\rlap{\begin{minipage}[t]{0.29\linewidth}
				\centering
				\begin{overpic}[width=0.879\textwidth]%
					{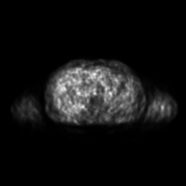}
					\centering
					\put(23,73){Input}
				\end{overpic}	
			\end{minipage}%
			\begin{minipage}[t]{1.41\linewidth}
				\centering
				\begin{overpic}[width=0.1815\textwidth]%
					{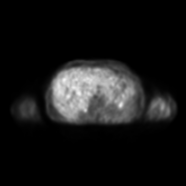}
					\centering
					\put(20,73){pix2pix}
				\end{overpic}
				\begin{overpic}[width=0.1815\textwidth]%
					{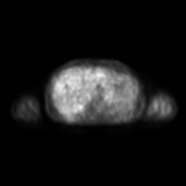}
					\centering
					\put(22,73){PAN}
				\end{overpic}
				\begin{overpic}[width=0.1815\textwidth]%
					{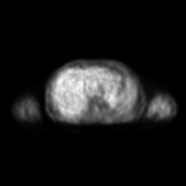}
					\centering
					\put(10,73){ID-CGAN}
				\end{overpic}
				\begin{overpic}[width=0.1815\textwidth]%
					{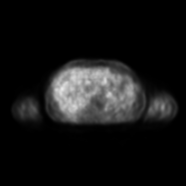}
					\centering
					\put(10,73){Fila-sGAN}
				\end{overpic}
				\begin{overpic}[width=0.1815\textwidth]%
					{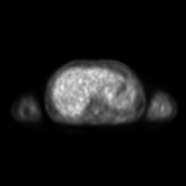}
					\centering
					\put(14,73){MedGAN}
				\end{overpic}
			\end{minipage}
			\begin{minipage}[t]{0.26\linewidth}
				\centering
				\begin{overpic}[width=0.978\textwidth]%
					{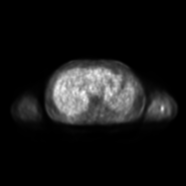}
					\centering
					\put(18,73){Target}
				\end{overpic}		
		\end{minipage}}
		\vspace{4mm}
		\rlap{\begin{minipage}[t]{0.29\linewidth}
				\centering
				\begin{overpic}[width=0.879\textwidth]%
					{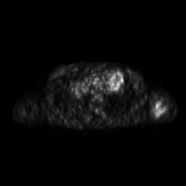}
					\centering
				\end{overpic}	
			\end{minipage}%
			\begin{minipage}[t]{1.41\linewidth}
				\centering
				\begin{overpic}[width=0.1815\textwidth]%
					{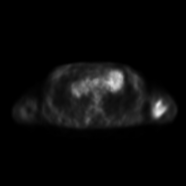}
					\centering
				\end{overpic}
				\begin{overpic}[width=0.1815\textwidth]%
					{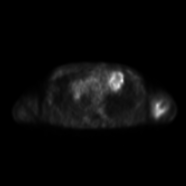}
					\centering
				\end{overpic}
				\begin{overpic}[width=0.1815\textwidth]%
					{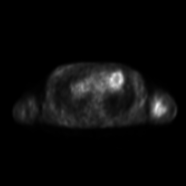}
					\centering
				\end{overpic}
				\begin{overpic}[width=0.1815\textwidth]%
					{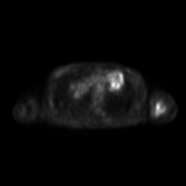}
					\centering
				\end{overpic}
				\begin{overpic}[width=0.1815\textwidth]%
					{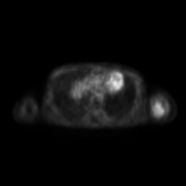}
					\centering
				\end{overpic}
			\end{minipage}
			\begin{minipage}[t]{0.26\linewidth}
				\centering
				\begin{overpic}[width=0.978\textwidth]%
					{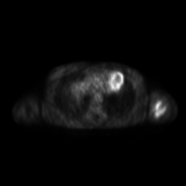}
					\centering
				\end{overpic}		
		\end{minipage}}
		
		\hspace{0.72\linewidth}\rlap{(c) PET denoising} 
	\end{minipage}

	\captionof{figure}{Additional results for MedGAN and other translat\rlap{ion approaches on the proposed medical translation tasks.}}
\end{center}


%
%
%
%
%




\end{document}